\documentclass{article}

\usepackage{microtype}
\usepackage{graphicx}
\usepackage{subcaption}
\usepackage{booktabs} 
\usepackage{bm}
\usepackage{multirow}
\usepackage{makecell}
\usepackage{amsmath}
\usepackage{placeins}
\usepackage{soul}
\usepackage{soulpos}

\usepackage{hyperref}

\newcommand{\Comment}[1]{\hfill \textit{// #1}}
\usepackage[table]{xcolor}
\definecolor{lightgray}{gray}{0.94} 
\newcommand{\g}[1]{\cellcolor{lightgray}#1}

\usepackage[preprint]{icml2026}

\usepackage{amsmath}
\usepackage{amssymb}
\usepackage{mathtools}
\usepackage{amsthm}

\usepackage[capitalize,noabbrev]{cleveref}

\theoremstyle{plain}

\theoremstyle{definition}

\theoremstyle{remark}

\usepackage[textsize=tiny]{todonotes}

\icmltitlerunning{LLM4Fluid: Large Language Models as Generalizable Neural Solvers for Fluid Dynamics}

\begin{document}

\twocolumn[
  \icmltitle{LLM4Fluid: Large Language Models as Generalizable Neural Solvers for Fluid Dynamics}



  \icmlsetsymbol{corr}{*}

  \begin{icmlauthorlist}
    \icmlauthor{Qisong Xiao}{1,2,3}
    \icmlauthor{Xinhai Chen}{corr,1,2,3}
    \icmlauthor{Qinglin Wang}{1,2,3}
    \icmlauthor{Xiaowei Guo}{1,2,3}
    \icmlauthor{Binglin Wang}{1,2,3}
    \icmlauthor{Weifeng Chen}{1,2,3}
    \icmlauthor{Zhichao Wang}{1,2,3}
    \icmlauthor{Yunfei Liu}{1,2,3}
    \icmlauthor{Rui Xia}{1,2,3}
    \icmlauthor{Hang Zou}{1,2,3}
    \icmlauthor{Gencheng Liu}{1,2,3}
    \icmlauthor{Shuai Li}{1,2,3}
    \icmlauthor{Jie Liu}{1,2,3}
  \end{icmlauthorlist}

  \icmlaffiliation{1}{National Key Laboratory of Parallel and Distributed Computing, National University of Defense Technology, Changsha 410073, China}
  \icmlaffiliation{2}{Laboratory of Digitizing Software for Frontier Equipment, National University of Defense Technology, Changsha 410073, China}
  \icmlaffiliation{3}{College of Computer Science and Technology, National University of Defense Technology, Changsha 410073, China}

  \icmlcorrespondingauthor{Xinhai Chen}{chenxinhai16@nudt.edu.cn}

  \icmlkeywords{Machine Learning, ICML}

  \vskip 0.3in
]



\printAffiliationsAndNotice{}  

\begin{abstract}
Deep learning has emerged as a promising paradigm for spatio-temporal modeling of fluid dynamics. However, existing approaches often suffer from limited generalization to unseen flow conditions and typically require retraining when applied to new scenarios. In this paper, we present LLM4Fluid, a spatio-temporal prediction framework that leverages Large Language Models (LLMs) as generalizable neural solvers for fluid dynamics. The framework first compresses high-dimensional flow fields into a compact latent space via reduced-order modeling enhanced with a physics-informed disentanglement mechanism, effectively mitigating spatial feature entanglement while preserving essential flow structures. A pretrained LLM then serves as a temporal processor, autoregressively predicting the dynamics of physical sequences with time series prompts. To bridge the modality gap between prompts and physical sequences, which can otherwise degrade prediction accuracy, we propose a dedicated modality alignment strategy that resolves representational mismatch and stabilizes long-term prediction. Extensive experiments across diverse flow scenarios demonstrate that LLM4Fluid functions as a robust and generalizable neural solver without retraining, achieving state-of-the-art accuracy while exhibiting powerful zero-shot and in-context learning capabilities. Code and datasets are publicly available at \href{https://github.com/qisongxiao/LLM4Fluid}{https://github.com/qisongxiao/LLM4Fluid}.
\end{abstract}

\section{Introduction}
\label{sec:introduction}

\begin{figure}[t]
	\centering
	\includegraphics[width=1.0\linewidth]{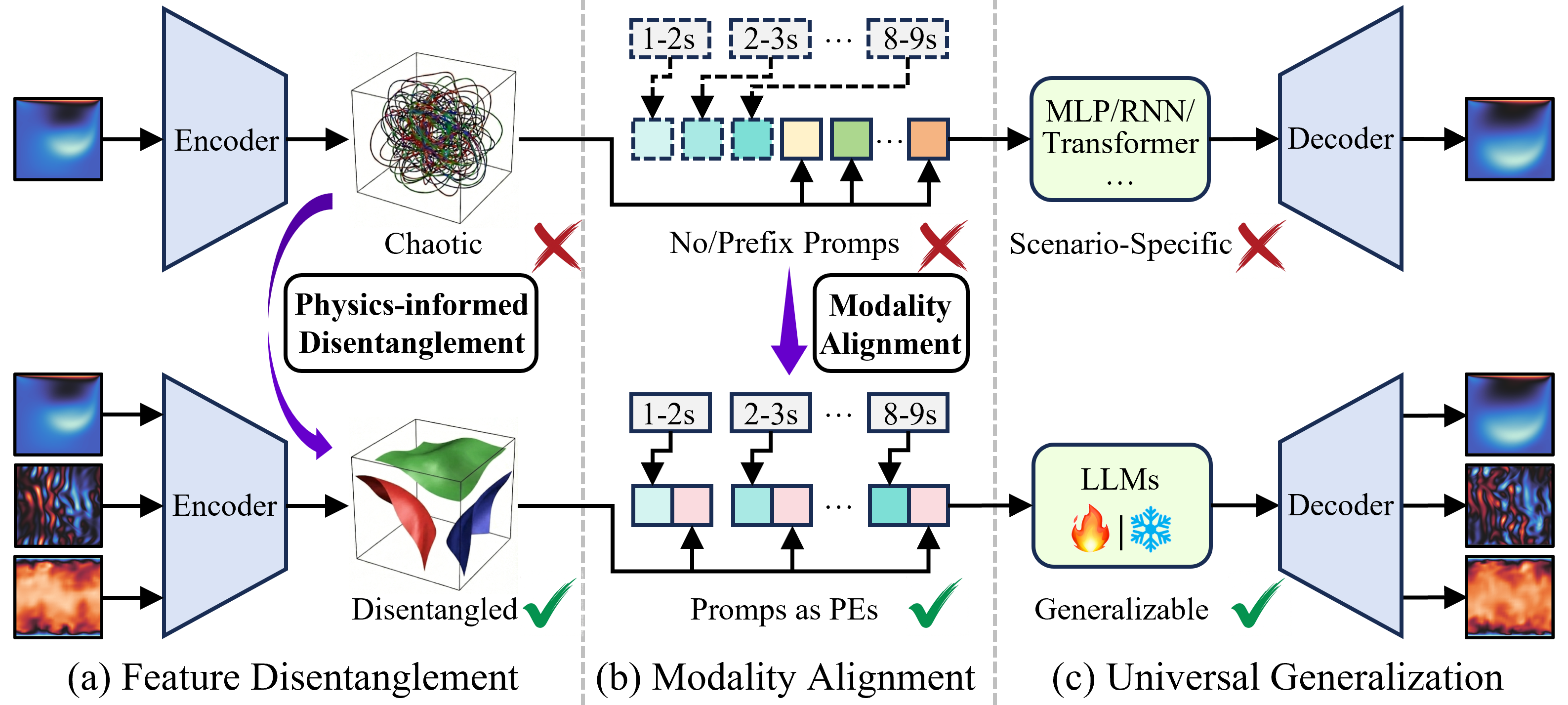}

	\caption{Comparison between existing methods (top) and our generalizable LLM4Fluid framework (bottom). (a) \textbf{Feature Disentanglement}: Unlike standard encoder–decoder architectures that produce entangled and chaotic spatial features in the latent space, our physics-informed disentanglement mechanism yields near-orthogonal and physics-disentangled representations while preserving essential flow structures. (b) \textbf{Modality Alignment}: Instead of prefix prompting, we embed textual prompts as position embeddings (PEs) to align the semantic and physical representations, bridging the modality gap and preventing prediction degradation. (c) \textbf{Universal Generalization}: Empowered by these mechanisms and pretrained sequence priors of LLMs, LLM4Fluid overcomes the limitations of scenario-specific methods, achieving robust generalization across diverse flow scenarios.}
	\label{fig:1}
\end{figure}

Fluid dynamics is fundamental to aerospace, ocean engineering, energy, and numerous scientific and engineering applications \cite{chen2024towards}. The inherently high-dimensional, nonlinear, and spatio-temporal nature of fluid motion poses severe challenges for accurate simulation, particularly in turbulent flows \cite{kim2024early}. Traditional computational fluid dynamics (CFD) solvers simulate spatio-temporal flow fields through numerical iterations, which suffer from prohibitive computational cost and slow convergence, becoming a major bottleneck for efficient simulation. \cite{vinuesa2022enhancing}.

Deep learning has emerged as a promising paradigm for fluid dynamics modeling, demonstrating accurate and efficient advantages across diverse CFD tasks \cite{peng20253dmeshnet,wang2025ugm2n,xiao2025st,shen2025peinr,xu2025amr}. However, spatio-temporal modeling of complex flow fields remains challenging due to the curse of dimensionality. Reduced-order models (ROMs) alleviate this problem by mapping flow fields into a low-dimensional space, yet traditional ROMs fail for nonlinear systems because of linear basis assumptions \cite{romor2025explicable}. Pioneer works have employed neural network-based ROMs such as multilayer perceptrons (MLPs) and convolutional neural networks (CNNs) to effectively compress flow fields into compact latent representations through encoder-decoder architectures \cite{xiao2024mh,mufti2025nonlinear}. Nevertheless, these approaches suffer from a spatial feature entanglement problem: the encoded spatial features are chaotic and mutually interfering due to the lack of explicit constraints, yielding entangled and redundant representations that fail to preserve physically meaningful flow structures, as illustrated in \cref{fig:1}. This problem destabilizes the latent space and severely impedes model generalization.

Furthermore, while temporal evolution is critical to fluid dynamics, existing methods focus solely on spatial compression and neglect temporal dependencies, restricting applicability to transient flows. To address this limitation, some approaches incorporate Transformers or recurrent neural networks (RNNs) as temporal processors \cite{solera2024beta,luo2026hybrid}. However, these methods are scenario-specific, with learned temporal dynamics tightly coupled to the training distribution, leading to poor generalization across varying flow scenarios. Consequently, tedious retraining is required to achieve reliable out-of-distribution predictions.

Recently, foundation models, particularly large language models (LLMs), have exhibited remarkable generalization and transferability in natural language processing and computer vision \cite{li2025fundamental}. Benefiting from large-scale pretraining, LLMs can generalize to new scenarios in few-shot or even zero-shot settings, and their inherent in-context learning capability enables knowledge transfer across diverse scenarios. Inspired by these strengths, researchers have explored the employment of LLMs for fluid dynamics \cite{zhu2024fluid,zou2025flowbert,wang2025can}, but most focus on modeling the physical sequence and ignore the semantic modality: textual-style prompts that describe the sequence context, such as the time range and sampling interval. Without the semantic modality, the model must implicitly infer temporal information from physical sequence, which weakens its ability to exploit pretrained sequence priors and often leads to unstable autoregressive rollout. Existing prompt-based methods typically treat prompts as prefixes concatenated with the physical sequence, neglecting the fundamental modality gap between semantic and physical representations, as illustrated in \cref{fig:1}. This representational mismatch impedes effective guidance of physical sequence evolution by prompts, amplifies cumulative errors, and ultimately leads to degraded long-term prediction stability.

To overcome the aforementioned limitations, we present an LLM-based neural solver for flow field prediction that achieves high accuracy, robust generalization, and stable long-term prediction. The main contributions are summarized as follows:

\begin{itemize}
	\item \textbf{Framework:} We propose LLM4Fluid, a spatio-temporal prediction framework that compresses high-dimensional flow fields into a compact latent space. A pretrained LLM then serves as a temporal processor to autoregressively predict the latent dynamics, enabling accurate and generalizable spatio-temporal modeling.
	\item \textbf{Method:} We develop a physics-informed disentanglement mechanism that yields near-orthogonal and physics-disentangled representations to mitigate spatial feature entanglement, and a modality alignment strategy that bridges the gap between semantic prompts and physical sequences to prevent prediction degradation.
	\item \textbf{Benchmark:} We construct a comprehensive benchmark tailored for fluid modeling, covering diverse flow scenarios with different boundary conditions, fluid properties, and domain geometries.
	\item \textbf{Experiments:} We evaluate LLM4Fluid with varying flow scenarios. Experimental results demonstrate that LLM4Fluid outperforms existing state-of-the-art methods in accuracy with minimal trainable parameters. It performs well in zero-shot and in-context prediction, demonstrating impressive cross-scenario generalization capability.
\end{itemize}

\section{Related Work}
\label{sec:related}

\subsection{Reduced-Order Models}

Reduced-order models (ROMs) alleviate the curse of dimensionality by mapping flow fields into a low-dimensional space. Traditional ROMs such as proper orthogonal decomposition (POD) \cite{jiang2025proper} and dynamic mode decomposition (DMD) \cite{lin2025enhanced} are widely used, but their linear basis assumptions limit their applicability to highly nonlinear physical systems. With the development of deep learning, neural-network-based ROMs have significantly improved nonlinear modeling capacity \cite{fu2025parametric}. Encoder–decoder architectures have become the mainstream framework for learning latent flow representations \cite{mallick2025ai}. CNN-based designs, such as convolutional autoencoders (CAEs) and U-Nets, extract spatial features hierarchically and achieve superior reconstruction fidelity compared to traditional ROMs \cite{vinograd2025reduced,grimm2025learning}. Despite these advances, most intelligent ROMs lack explicit constraints in the latent space, leading to spatial feature entanglement and undermining physical interpretability. This further results in unstable latent dynamics and degraded prediction performance.

\subsection{Temporal Processors}

Incorporating a temporal processor between the encoder and decoder has been proven to be an effective strategy for modeling the temporal evolution of latent representations \cite{wang2025can}. Existing studies utilize a wide range of processors, including MLP-based \cite{wang2025filterts}, RNN-based \cite{beck2024xlstm}, CNN-based \cite{cheng2025convtimenet}, and Transformer-based architectures \cite{liu2024itransformer}. More recent advances investigate new temporal modeling paradigms, such as state space models \cite{dao2024transformers} and Kolmogorov–Arnold Networks (KANs) \cite{huang2025timekan}, which have demonstrated promising and competitive performance on sequence prediction tasks. However, these approaches remain scenario-specific and require retraining once flow conditions change. To overcome this limitation, recent studies have explored LLM-based processors, which can be broadly categorized into time series-based and prompt-based frameworks \cite{liu2025timecma}. The former replaces the tokenizer with a randomly initialized embedding layer to directly encode time series data, as in GPT4TS \cite{zhou2023one}, whereas the latter incorporates prompts as additional inputs to assist prediction, such as Time-LLM \cite{jin2024timellm}. Nevertheless, the above methods still fail to prevent prediction degradation induced by the inherent modality gap between semantic and physical representations.

\section{Methodology}
\label{sec:methodology}

\begin{figure*}[t]
	\centering
	\includegraphics[width=0.9\linewidth]{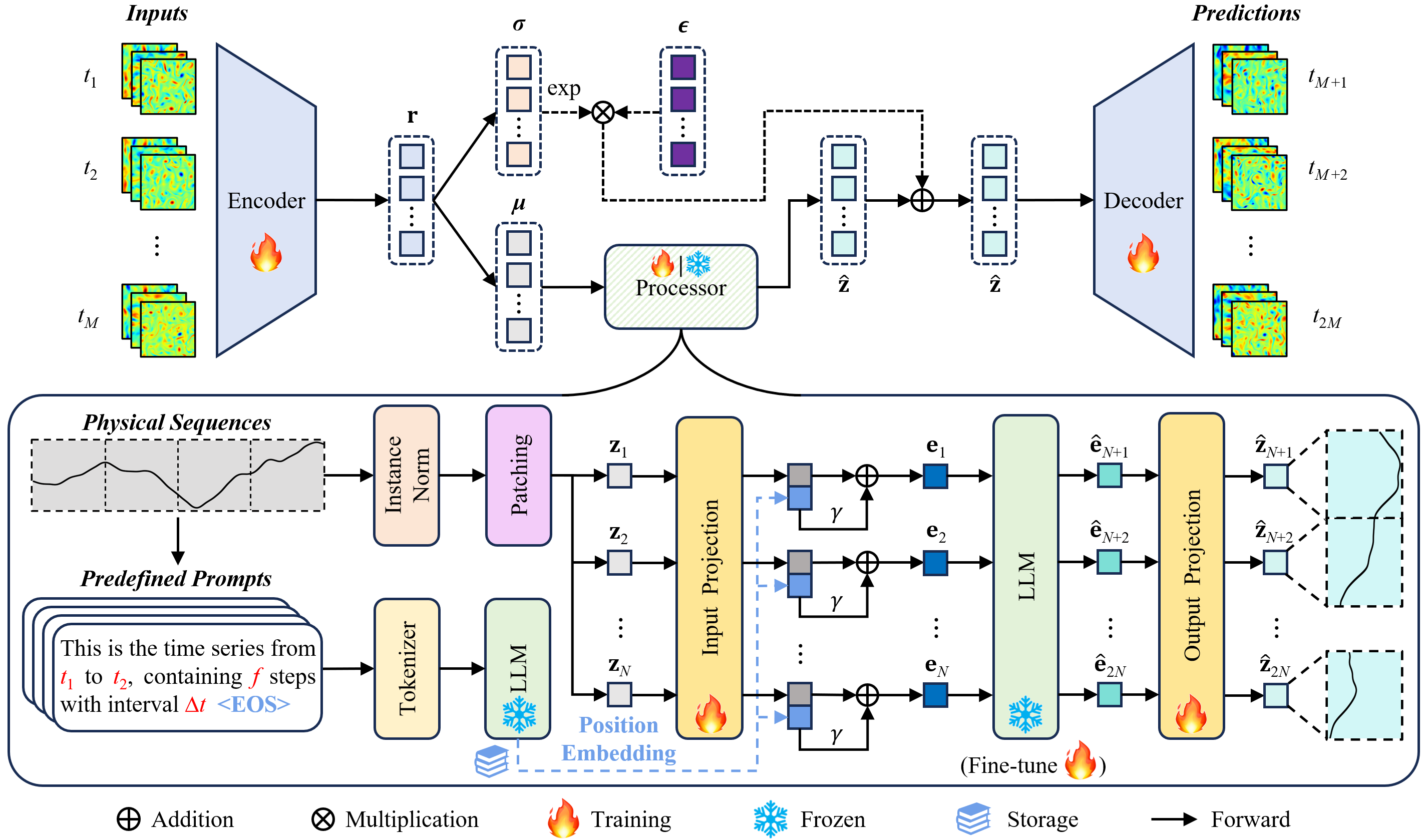}
	\vskip -0.1cm
	\caption{Overall architecture of the proposed LLM4Fluid framework. LLM4Fluid operates in two stages: (1) disentangled reduced-order modeling that compresses high-dimensional flow fields into near-orthogonal and physics-disentangled latent representations; and (2) an LLM-based temporal processor that tokenizes physical sequences, incorporates predefined prompts as positional embeddings, and autoregressively predicts future latent states. The predicted physical sequences are finally decoded to reconstruct the future flow fields.}
	\vskip -0.2cm
	\label{fig:2}
\end{figure*}

\begin{figure}[t]
	\centering
	\includegraphics[width=1.0\linewidth]{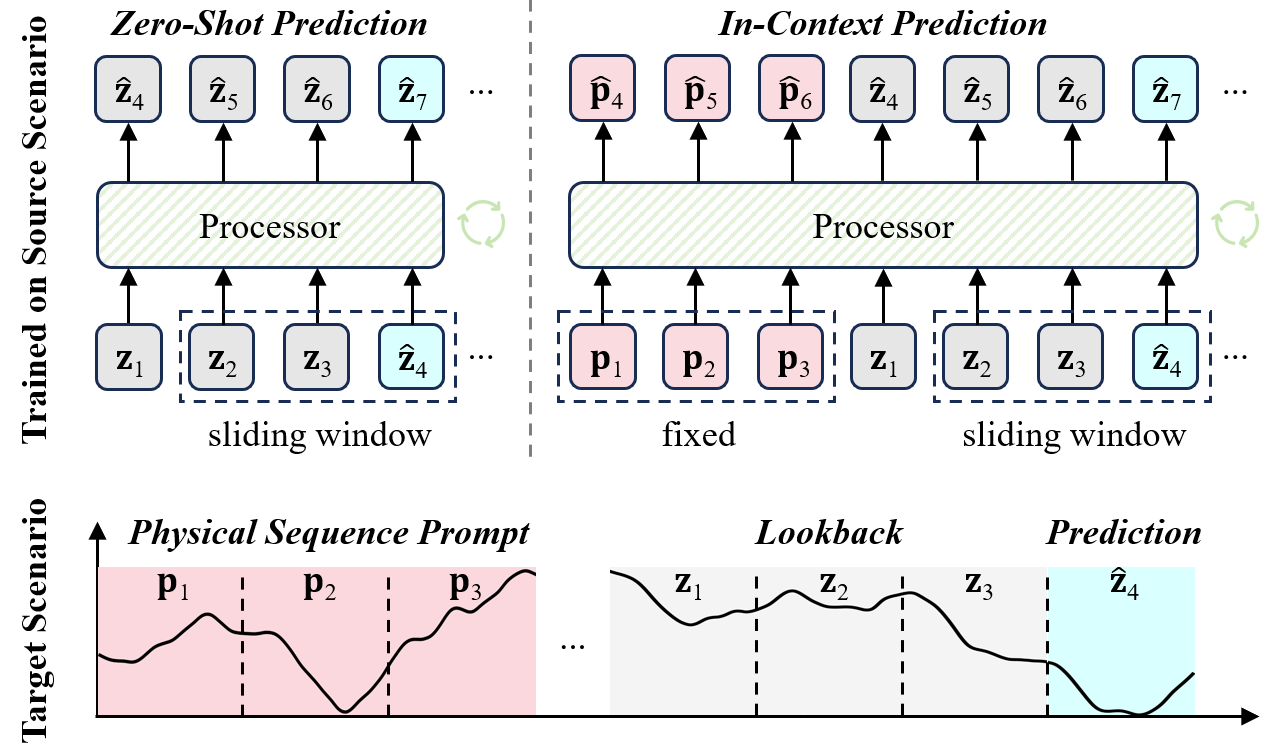}
	\vskip -0.1cm
	\caption{Overview of zero-shot and in-context prediction. Trained on a source scenario, the processor performs autoregressive prediction on a target scenario either using only the lookback sequence (zero-shot) or additionally with physical sequence prompts (in-context).}
	\label{fig:3}
	\vskip -0.2cm
\end{figure}

\subsection{Task Definition} \label{task definition}

Fluid dynamics exhibits complex, high-dimensional, and nonlinear flow behaviors, and accurately predicting its spatio-temporal evolution is fundamental for uncovering the underlying physical mechanism. Given a sequence of flow field snapshots $\mathbf{X}_{1:T}=\left\{\mathbf{X}_1, \ldots, \mathbf{X}_T\right\}$, where $\mathbf{X}_t \in \mathbb{R}^{H \times W \times C} (t=1,...,T)$. $H$ and $W$ denote the spatial resolution and $C$ represents the number of physical variables, the objective is to predict the next $F$ flow fields. The prediction task aims to train a spatio-temporal predictor $f$ that satisfies
\begin{equation}
	f: \mathbf{X}_{1: T} \mapsto \widehat{\mathbf{X}}_{T+1: T+F}.
	\label{eq:1}
\end{equation}

\subsection{Prediction Framework} \label{prediction framework}

The proposed LLM4Fluid framework is designed to achieve accurate, efficient, and generalizable long-term spatio-temporal prediction of fluid dynamics. As illustrated in \cref{fig:2}, the overall framework consists of two sequential training stages. In the first stage, disentangled reduced-order modeling is performed. High-dimensional flow fields $\mathbf{X}_t$ are compressed through an encoder–decoder architecture enhanced with a physics-informed disentanglement mechanism, ensuring the preservation of physically meaningful flow structures. After training, this module provides near-orthogonal and physics-disentangled latent representations $\mathbf{z}_t \in \mathbb{R}^D$, where $D$ denotes the dimension of the latent space. In the second stage, the above representations are arranged into physical sequences $\mathbf{z}_{1:T}=\left\{\mathbf{z}_1, \ldots, \mathbf{z}_T\right\}$, reflecting their temporal evolution, and subsequently partitioned into patches. These patched sequences, together with their corresponding prompts serving as positional embeddings, are fed into an LLM-based temporal processor, which autoregressively predicts future latent states. Finally, the predicted physical sequences are passed through the pretrained decoder from the first stage to reconstruct the future flow fields $\widehat{\mathbf{X}}_{T+1: T+F}$. The corresponding detailed algorithms are provided in \cref{sec:implementation details}.

\subsection{Disentangled Reduced-Order Modeling} \label{disentangled rom}

\noindent\textbf{Latent space construction.} Given an input flow field $\mathbf{X}$, the encoder $\mathcal{E}$ maps it to a latent representation
$\mathbf{r}=\mathcal{E}(\mathbf{X})$, and the decoder $\mathcal{D}$ reconstructs the field as
$\widehat{\mathbf{X}} = \mathcal{D}(\mathbf{r})$.
The latent space is learned by minimizing the reconstruction loss
\begin{equation}
	\mathcal{L}_{\mathrm{rec}}=\frac{1}{H\times W} \sum_{h=1}^H \sum_{w=1}^W\|\mathbf{X}^{(h,w)}-\widehat{\mathbf{X}}^{(h,w)}\|^2_2,
	\label{eq:2}
\end{equation}
where $\mathbf{X}^{(h,w)}\in\mathbb{R}^C$ denotes the physical variables at the spatial mesh point $(h,w)$ and $H\times W$ denotes the total number of points. However, the lack of explicit constraints causes the latent space to be contaminated with chaotic and redundant features unrelated to the underlying fluid dynamics, resulting in entangled representations that hinder model generalization and degrade long-term prediction stability.

\noindent\textbf{Physics-informed disentanglement mechanism.} To mitigate spatial feature entanglement, we develop a physics-informed disentanglement mechanism that imposes physical regularization on the latent representations. Specifically, the latent representation $\mathbf{r}$ is further projected through two linear heads to obtain the mean $\boldsymbol{\mu}$ and the standard deviation $\boldsymbol{\sigma}$. A latent sample is then generated via the reparameterization trick $\mathbf{z}=\boldsymbol{\mu}+\boldsymbol{\sigma} \odot \boldsymbol{\epsilon}$, where $\mathbf{z} \sim \mathcal{N}(\boldsymbol{\mu},\boldsymbol{\sigma})$, and $\odot$ denotes element-wise multiplication \cite{kingma2022auto}. The sampled latent variable is subsequently decoded back into the physical space as $\widehat{\mathbf{X}} = \mathcal{D}(\mathbf{z})$.

To promote the latent representations to be physics-disentangled, we incorporate the reconstruction objective with a physics-informed disentanglement loss term weighted by $\lambda$, which governs the trade-off between reconstruction fidelity and the degree of disentanglement. The complete optimization objective is
\begin{equation}
	\mathcal{L}=\mathcal{L}_{\mathrm{rec}}\underbrace{-\frac{\lambda}{2} \sum_{k=1}^D\left(1+\log \left(\sigma_k^2\right)-\mu_k^2-\sigma_k^2\right)}_{\text {Disentanglement loss }},
	\label{eq:3}
\end{equation}
where $\mu_k$ and $\sigma_k$ denote the $k$-th components of $\boldsymbol{\mu}$ and $\boldsymbol{\sigma}$, respectively.
The physics-informed disentanglement loss penalizes correlations across latent dimensions and encourages each axis to represent statistically independent modes of flow variation. This suppresses redundant interactions and yields a latent space that is near-orthogonal, smooth, and physics-disentangled, thereby providing a stable dynamical coordinate system that substantially benefits downstream temporal modeling and enhances robustness in long-term prediction. During temporal modeling, we use $\mathbf{z}=\boldsymbol{\mu}$ to avoid sampling noise and ensure stable propagation.

\subsection{LLM-based Temporal Processor} \label{llm-based temporal processor}

\noindent\textbf{Physical sequence tokenization.} 
We perform prediction on a sliding window of length $M$. Given the physical sequence $\mathbf{z} \in \mathbb{R}^{D\times M}$, each latent variable $\mathbf{z}^{(i)} \in \mathbb{R}^{M}$ ($i=1,\dots,D$) is modeled independently to avoid mutual interference. We first apply reversible instance normalization (RevIN) to alleviate temporal distribution shift \cite{kim2021reversible}. Then, $\mathbf{z}^{(i)}$ is divided into non-overlapping patches of length $M_p$, yielding $N = \left\lfloor M/M_p \right\rfloor$ patches. The patching operation preserves local physical information while serving as a tokenization mechanism, producing a compact physical token sequence $\mathbf{z}^{(i)}_{1:N}=\{{\mathbf{z}^{(i)}_1, …, \mathbf{z}^{(i)}_N}\}$. An MLP-based input projection layer with Mish activation function subsequently processes these tokens:
\begin{equation}
	\mathbf{s}^{(i)}_{1:N}=\mathrm{InputProjection}(\mathbf{z}^{(i)}_{1:N}),
	\label{eq:4}
\end{equation}
where $\mathbf{s}^{(i)}_{1:N} \in \mathbb{R}^{D_e \times N}$ denotes the physical embeddings and $D_e$ denotes the embedding dimension. Although each latent variable is processed independently, they share the same model parameters, enabling the network to implicitly capture cross-variable correlations.

\noindent\textbf{Modality alignment.}
Textual prompts have been demonstrated to effectively improve temporal modeling performance, commonly achieved through prefix prompting \cite{jin2024timellm,zou2025flowbert}. However, directly concatenating textual prompts with physical sequence often leads to a modality mismatch. To address this problem, we explicitly align the two modalities by employing LLM-embedded prompts as position embeddings.

Given a fixed window size $M$ and patch length $M_p$, we predefine a sequence of textual prompts $\mathrm{Text}_{1:N}$ corresponding to the $N$ physical patches, as illustrated in \cref{fig:2}. These prompts are tokenized and processed by the frozen LLM, and left-padding is applied to maintain a consistent token length. To obtain semantic representations aligned with the physical embeddings, we extract the last special token \texttt{<EOS>} from the LLM output, which contains the overall semantic meaning of each prompt. The LLM-embedded prompts are used as position embeddings and fused with the physical embeddings:
\begin{gather}
	\mathbf{k}_{1:N}=\mathrm{Last}(\mathrm{LLM}(\mathrm{Tokenizer}(\mathrm{Text}_{1:N}))),
	\label{eq:5} \\
	\mathbf{e}^{(i)}_{1:N}= \mathbf{s}^{(i)}_{1:N}+\gamma\cdot\mathbf{k}_{1:N},
	\label{eq:6}
\end{gather}
where $\mathbf{e}^{(i)}_{1:N}$ denotes the physical embeddings enriched with textual prompt information, and $\gamma$ is a learnable scaling factor that adjusts the contributions of the prompt embeddings.

\noindent\textbf{Autoregressive prediction.}
LLMs possess powerful sequence modeling and reasoning capabilities, enabling them to autoregressively predict future embeddings based on preceding ones. We reuse this autoregressive mechanism in our framework and incorporate a sliding window strategy, allowing the model to generate arbitrarily long sequences under a fixed attention cost. The aligned physical embeddings $\mathbf{e}^{(i)}_{1:N}$ are fed into the frozen LLM to predict the embeddings at future time steps:
\begin{equation}
	\widehat{\mathbf{e}}^{(i)}_{N+1:2N}=\mathrm{LLM}(\mathbf{e}^{(i)}_{1:N}).
	\label{eq:7}
\end{equation}

Finally, we adopt $\mathrm{OutputProjection}(\cdot): \mathbb{R}^{D_e} \mapsto \mathbb{R}^D$ to map the predicted embeddings back to the latent space:
\begin{equation}
	\widehat{\mathbf{z}}^{(i)}_{N+1:2N}=\mathrm{OutputProjection}(\widehat{\mathbf{e}}^{(i)}_{N+1:2N}).
	\label{eq:8}
\end{equation}

The input and output projection layers are optimized using the mean squared error (MSE) loss:
\begin{equation}
	\mathcal{L_\mathrm{mse}}=\frac{1}{D\times N} \sum_{i=1}^D \sum_{j=N+1}^{2N}\| \mathbf{z}^{(i)}_j-\widehat{\mathbf{z}}^{(i)}_j\|^2_2.
	\label{eq:9}
\end{equation}

The LLM-based temporal solver generates the target latent representations $\widehat{\mathbf{z}}_{T+1:T+F}$ through autoregressive prediction. These representations are then passed through the decoder trained in the disentangled reduced-order modeling stage to obtain the reconstructed flow fields $\widehat{\mathbf{X}}_{T+1:T+F}$.

\begin{figure}[t]
	\centering
	\includegraphics[width=1.0\linewidth]{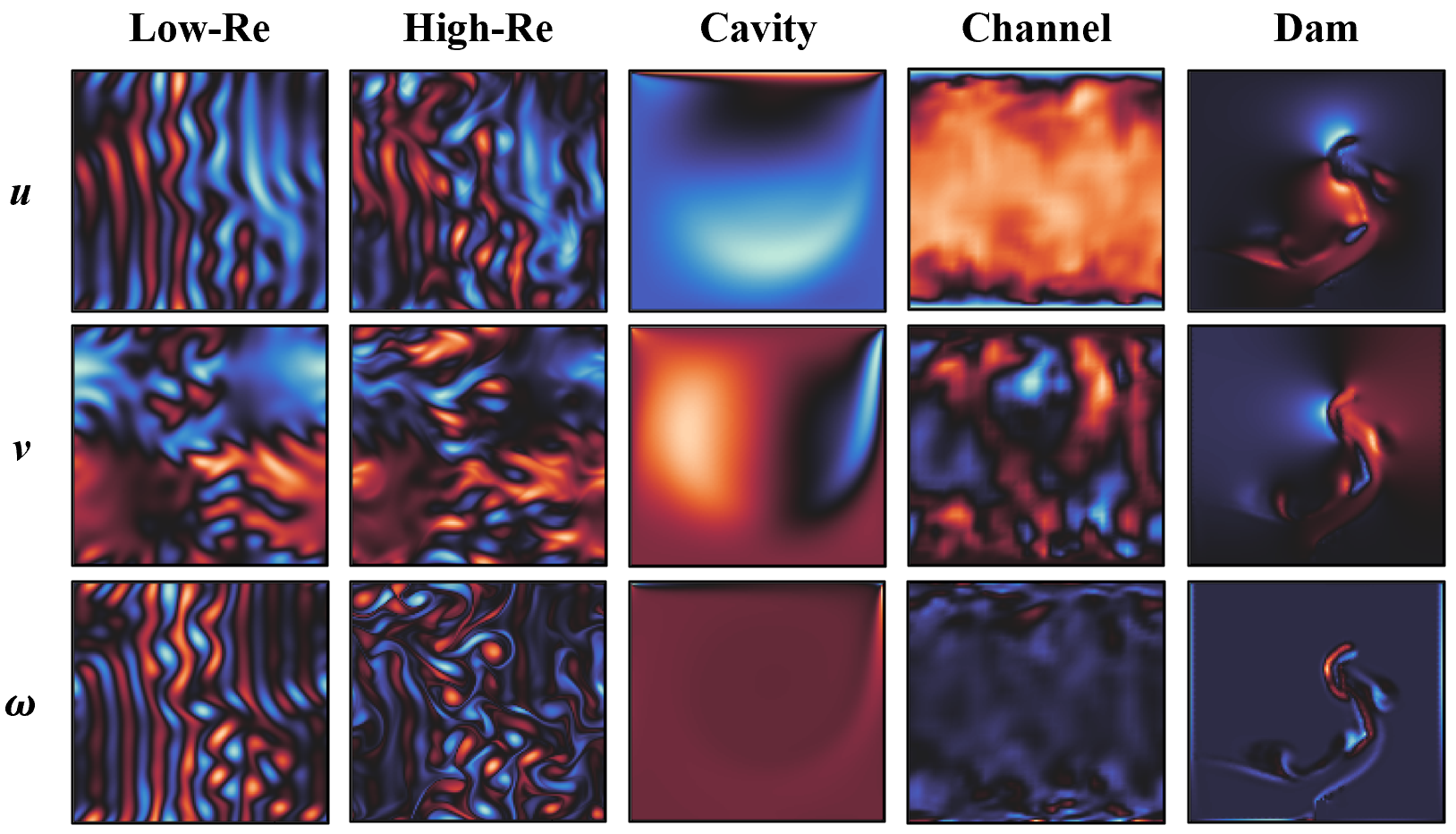}
	\vskip -0.1cm
	\caption{Visualization of flow fields across five datasets.}
	\vskip -0.2cm
	\label{fig:4}
\end{figure}

\noindent\textbf{In-context learning.}
LLMs exhibit in-context learning (ICL) capabilities, enabling them to rapidly adapt to new scenarios through demonstration prompts without requiring parameter updates. We exploit this property to enhance the generalization capability of LLM4Fluid, allowing the pretrained LLM to infer the temporal evolution mechanism of the underlying physical system conditioned on a context set $\mathcal{C}$. Let $\{\mathbf{p}^{(i)}_1, \dots, \mathbf{p}^{(i)}_{2n}\}$ ($i=1,\dots,D$) denote the physical patches obtained from a continuous physical sequence in the target scenario. The context set, which consists of $n$ pairs of preceding and subsequent physical patches, is defined as:
\begin{equation}
	\mathcal{C}=\{{(\mathbf{p}^{(i)}_1, \mathbf{p}^{(i)}_{n+1}), …, (\mathbf{p}^{(i)}_n, \mathbf{p}^{(i)}_{2n})}\},
	\label{eq:10}
\end{equation}
where each $\mathbf{p}^{(i)}_j \in \mathbb{R}^{M_p}$ ($j=1,\dots,2n$).
During prediction, the processor receives both the physical patches within the lookback window and the context set sampled from the target scenario. The context set $\mathcal{C}$ remain fixed, whereas the lookback sequence is autoregressively updated through the sliding window strategy, as illustrated in \cref{fig:3}.

\section{Experiments}
\label{sec:experiment}

\subsection{Experimental Setups} \label{sec:experimental setups}
\noindent\textbf{Datasets.}
We evaluate all models on five datasets covering diverse flow scenarios, including Kolmogorov flow, lid-driven cavity flow (Cavity), channel flow (Channel), and dam-break flow (Dam), as illustrated in \cref{fig:4}. Specifically, Low-Re and High-Re denote Kolmogorov flows with Reynolds numbers $\mathrm{Re}=100$ and $\mathrm{Re}=1000$, respectively. Each dataset contains 1500 flow field snapshots at a spatial resolution of $128 \times 128$. The physical variables include the streamwise velocity ($u$), vertical velocity ($v$), and vorticity ($\omega$). Each dataset is split into training and test sets with a 9:1 ratio. More details are provided in \cref{sec:dataset details}.

\begin{table}[t]
	\centering
	\caption{Impact of disentanglement weight $\lambda$ on reconstruction performance on the High-Re dataset.}
	\label{tab:1}
	\scriptsize
	\setlength{\tabcolsep}{3pt}
	\vskip -0.1cm
	\resizebox{1.0\linewidth}{!}{
		\begin{tabular}{@{\extracolsep{\fill}}c|
				ccc|
				ccc}
			\toprule
			\multicolumn{1}{c}{\multirow{2}{*}{Weight}} &
			\multicolumn{3}{c}{Error} &
			\multicolumn{3}{c}{Quality} \\
			\cmidrule(lr){2-4} \cmidrule(lr){5-7}
			\multicolumn{1}{c}{~} &
			\multicolumn{1}{c}{MAE $\downarrow$} &
			\multicolumn{1}{c}{MSE $\downarrow$} &
			\multicolumn{1}{c}{SMAPE $\downarrow$} &
			\multicolumn{1}{c}{$\text{R}^\text{2}$ $\uparrow$} &
			\multicolumn{1}{c}{PSNR $\uparrow$} &
			\multicolumn{1}{c}{SSIM $\uparrow$} \\
			\midrule
			0     & 6.016E-2 & 7.714E-3 & 5.262E-1 & 0.889 & 21.379 & 0.673 \\
			1E-1  & 1.446E-1 & 3.745E-2 & 9.630E-1 & 0.459 & 14.274 & 0.235 \\
			1E-2  & 1.426E-1 & 3.696E-2 & 9.500E-1 & 0.467 & 14.331 & 0.259 \\
			1E-3  & 6.751E-2 & 9.491E-3 & 5.716E-1 & 0.863 & 20.479 & 0.619 \\
			1E-4  & \textbf{5.513E-2} & \underline{6.810E-3} & \textbf{4.914E-1} & \underline{0.902} & \underline{22.054} & \textbf{0.706} \\
			1E-5  & \underline{5.658E-2} & \textbf{6.750E-3} & \underline{5.144E-1} & \textbf{0.903} & \textbf{22.058} & \underline{0.684} \\
			1E-6  & 5.920E-2 & 7.587E-3 & 5.201E-1 & 0.890 & 21.575 & 0.675 \\
			\bottomrule
		\end{tabular}
	}

\end{table}

\begin{figure}[t]
	\centering
	\vskip -0.1cm
	\begin{subfigure}[b]{0.48\linewidth}
		\centering
		\includegraphics[width=\linewidth]{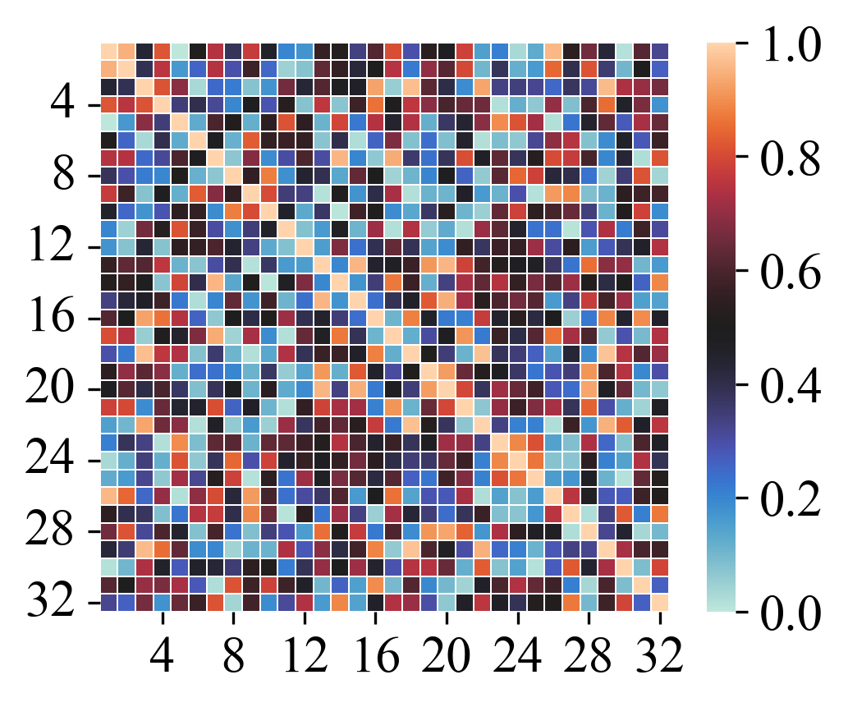}
		\subcaption{$\lambda$=0}
		\label{fig:5a}
	\end{subfigure}
	\hfill
	\begin{subfigure}[b]{0.48\linewidth}
		\centering
		\includegraphics[width=\linewidth]{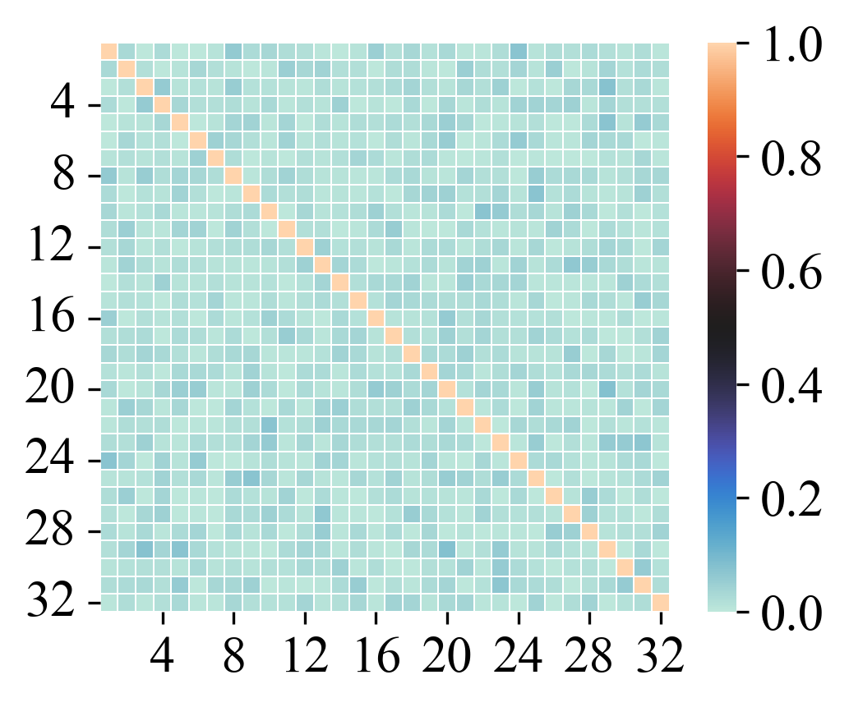}
		\subcaption{$\lambda$=1E-3}
		\label{fig:5b}
	\end{subfigure}
	
	\begin{subfigure}[b]{0.48\linewidth}
		\centering
		\includegraphics[width=\linewidth]{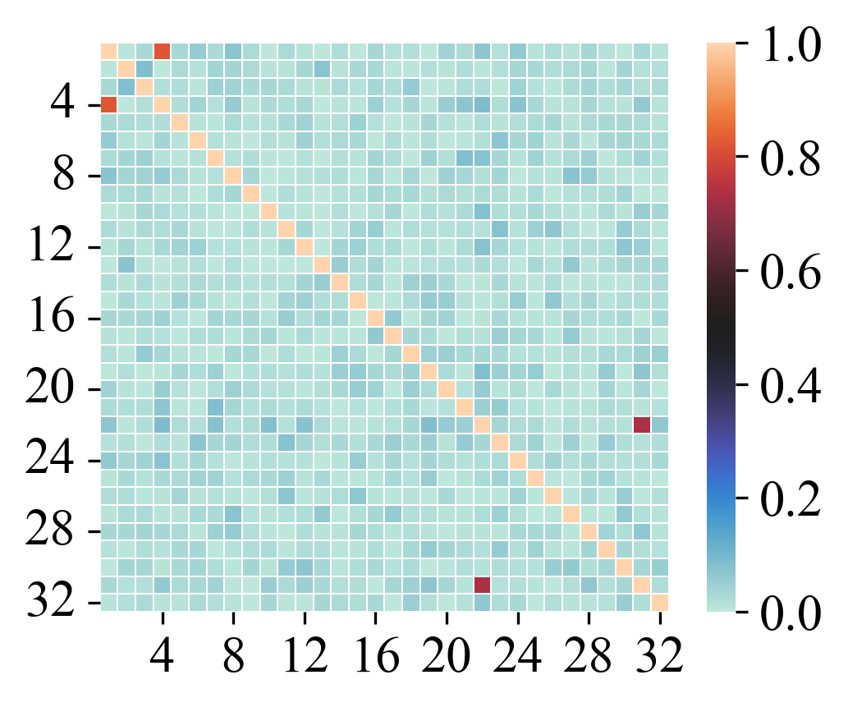}
		\subcaption{$\lambda$=1E-4}
		\label{fig:5c}
	\end{subfigure}
	\hfill
	\begin{subfigure}[b]{0.48\linewidth}
		\centering
		\includegraphics[width=\linewidth]{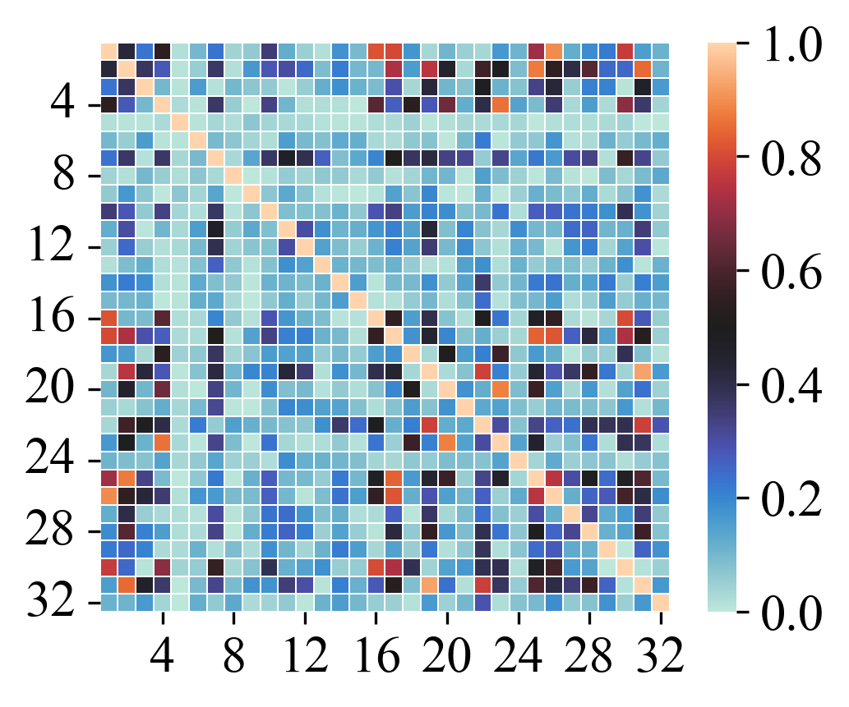}
		\subcaption{$\lambda$=1E-5}
		\label{fig:5d}
	\end{subfigure}
	
	\caption{Correlation matrices of the latent representations under different disentanglement weights on the High-Re dataset.}
	\label{fig:5}
	\vskip -0.2cm
\end{figure}

\begin{figure}[th]
	\centering
	\includegraphics[width=1.0\linewidth]{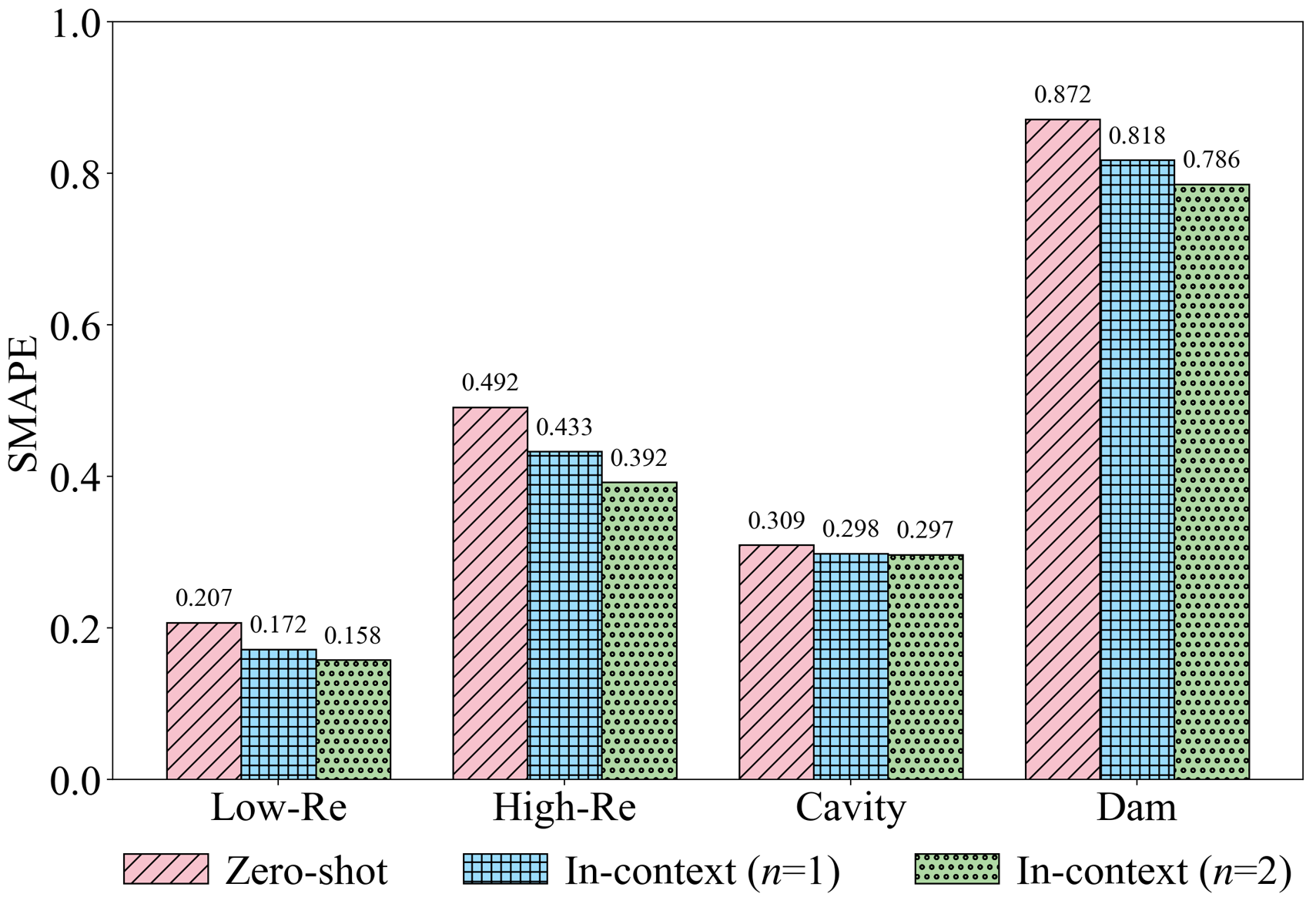}
	\vskip -0.1cm
	\caption{Comparison of SMAPE for LLM4Fluid in zero-shot and in-context prediction with different numbers of physical patch pairs $\textit{n}$ used as prompts. LLM4Fluid is trained on the Channel dataset and evaluated on the other datasets. In-context learning improves prediction performance across diverse flow scenarios.}
	\vskip -0.2cm
	\label{fig:6}
\end{figure}

\noindent\textbf{Baselines.}
We compare LLM4Fluid against a broad range of state-of-the-art (SOTA) time-series prediction models. MLP-based models include HDMixer \cite{huang2024hdmixer}, SOFTS \cite{han2024softs}, TimeMixer \cite{wang2024timemixer}, WPMixer \cite{murad2025wpmixer}, CrossLinear \cite{zhou2025crosslinear}, and FilterTS \cite{wang2025filterts}; RNN-based models include LSTM \cite{hou2022novel} and xLSTM \cite{beck2024xlstm}; CNN-based models include ConvTimeNet \cite{cheng2025convtimenet}; Transformer-based models include iTransformer \cite{liu2024itransformer}, TimeXer \cite{wang2024timexer}, CASA \cite{lee2025casa}, and TimeBridge \cite{liu2025timebridge}; Mamba-based models include Mamba \cite{gu2024mamba} and Mamba2 \cite{dao2024transformers}; KAN-based models include TimeKAN \cite{huang2025timekan}; LLM-based models include GPT4TS \cite{zhou2023one} and Time-LLM \cite{jin2024timellm}. To ensure a fair comparison, we adjust the number of trainable parameters to similar levels.

\noindent\textbf{Evaluation metrics.}
To comprehensively evaluate prediction performance, we report mean absolute error (MAE), mean squared error (MSE), and symmetric mean absolute percentage error (SMAPE) as error metrics. For quality metrics, we use the coefficient of determination ($\text{R}^\text{2}$), peak signal-to-noise ratio (PSNR), and structural similarity index measure (SSIM). Detailed descriptions of these metrics are provided in \cref{sec:evaluation metrics}.

\noindent\textbf{Implementation.}
For all experiments, we train models with AdamW using an initial learning rate of 1E-3 and a batch size of 128. All models are trained for 200 epochs on a single NVIDIA RTX 5090 GPU. We set the sliding window length to $M=20$ and the patch length to $M_p=5$, yielding $N=4$ patches per lookback window. We use OPT-6.7B as the LLM backbone and fine-tune it with LoRA \cite{hu2022lora} for each dataset. The latent dimension and embedding dimension are set to $D=32$ and $D_e=4096$, respectively. More details are provided in \cref{sec:implementation details}.

\begin{table*}[t]
	\centering
	\caption{Prediction performance of different models on five datasets. Trainable parameters are measured in megabytes (MB). The best and second-best performances are highlighted in \textbf{bold} and \underline{underlined}, respectively.}
	\label{tab:2}
	\scriptsize
	\setlength{\tabcolsep}{3pt}
	\vskip -0.1cm
	\resizebox{1.0\linewidth}{!}{
		\begin{tabular}{
				c|c|c|c|
				cc|cc|
				cc|cc|cc}
			\toprule
			\multicolumn{1}{c}{\multirow{2}{*}{\textbf{Type}}} &
			\multicolumn{1}{c}{\multirow{2}{*}{\textbf{Model}}} &
			\multicolumn{1}{c}{\multirow{2}{*}{\textbf{Source}}} &
			\multicolumn{1}{c}{\multirow{2}{*}{\makecell[c]{\textbf{Trainable}\\\textbf{Parameters}}}} &
			\multicolumn{2}{c}{\textbf{Low-Re}} &
			\multicolumn{2}{c}{\textbf{High-Re}} &
			\multicolumn{2}{c}{\textbf{Cavity}} &
			\multicolumn{2}{c}{\textbf{Channel}} &
			\multicolumn{2}{c}{\textbf{Dam}} \\
			\cmidrule(lr){5-6} \cmidrule(lr){7-8} \cmidrule(lr){9-10} \cmidrule(lr){11-12} \cmidrule(lr){13-14}
			\multicolumn{4}{c}{~} &
			\multicolumn{1}{c}{MAE $\downarrow$} & \multicolumn{1}{c}{MSE $\downarrow$} &
			\multicolumn{1}{c}{MAE $\downarrow$} & \multicolumn{1}{c}{MSE $\downarrow$} &
			\multicolumn{1}{c}{MAE $\downarrow$} & \multicolumn{1}{c}{MSE $\downarrow$} &
			\multicolumn{1}{c}{MAE $\downarrow$} & \multicolumn{1}{c}{MSE $\downarrow$} &
			\multicolumn{1}{c}{MAE $\downarrow$} & \multicolumn{1}{c}{MSE $\downarrow$} \\
			\midrule
			\multirow{4}{*}{Transformer} & iTransformer & \textit{ICLR 2024} & 1.550 & 1.404E-1 & 4.010E-2 & 3.024E-1 & 1.766E-1 & 2.173E-1 & 1.203E-1 & 4.944E-1 & 4.966E-1 & 1.569E-1 & 6.428E-2 \\
			& TimeXer & \textit{NeurIPS 2024} & 1.610 & 1.867E-2 & 6.846E-4 & 5.806E-2 & 7.598E-3 & 2.185E-2 & 1.885E-3 & 4.772E-2 & 6.673E-3 & 3.415E-2 & 6.229E-3 \\
			& CASA & \textit{IJCAI 2025} & 1.911 & 1.851E-2 & 6.800E-4 & 5.429E-2 & 6.521E-3 & 2.645E-2 & 3.691E-3 & 4.860E-2 & 6.963E-3 & 2.088E-2 & 1.923E-3 \\
			& TimeBridge & \textit{ICML 2025} & 1.704 & 1.854E-2 & 6.843E-4 & 5.414E-2 & 6.656E-3 & 9.423E-3 & 2.754E-4 & 2.950E-2 & 2.242E-3 & 2.924E-2 & 5.436E-3 \\
			\midrule
			\multirow{1}{*}{KAN} & TimeKAN & \textit{ICLR 2025} & 1.778 & 2.351E-2 & 1.161E-3 & 5.717E-2 & 7.371E-3 & 8.264E-3 & 1.587E-4 & 4.616E-2 & 6.037E-3 & 3.035E-2 & 5.128E-3 \\
			\midrule
			\multirow{6}{*}{MLP} & HDMixer & \textit{AAAI 2024} & 1.627 & 1.869E-2 & 6.883E-4 & 5.287E-2 & 6.248E-3 & 6.241E-3 & 8.432E-5 & 1.207E-1 & 5.883E-2 & 2.980E-2 & 4.970E-3 \\
			& SOFTS & \textit{NeurIPS 2024} & 1.658 & 1.854E-2 & 6.842E-4 & 5.628E-2 & 7.153E-3 & 7.169E-3 & 1.151E-4 & 3.126E-2 & 2.628E-3 & 2.764E-2 & 4.366E-3 \\
			& TimeMixer & \textit{ICLR 2024} & 1.717 & 1.879E-2 & 7.015E-4 & 5.659E-2 & 7.234E-3 & 8.092E-3 & 1.520E-4 & 4.532E-2 & 5.829E-3 & 3.005E-2 & 5.037E-3 \\
			& WPMixer & \textit{AAAI 2025} & 1.576 & 1.893E-2 & 7.022E-4 & 1.385E-1 & 4.610E-2 & 2.761E-2 & 5.846E-3 & 3.097E-2 & 2.374E-3 & 3.380E-2 & 5.589E-3 \\
			& CrossLinear & \textit{SIGKDD 2025} & 1.709 & 1.875E-2 & 6.897E-4 & 5.736E-2 & 7.292E-3 & 8.884E-3 & 2.335E-4 & 4.122E-2 & 4.891E-3 & 3.006E-2 & 5.054E-3 \\
			& FilterTS & \textit{AAAI 2025} & 1.691 & 1.947E-2 & 7.424E-4 & 5.280E-2 & 6.312E-3 & 9.939E-3 & 3.188E-4 & 7.019E-2 & 1.565E-2 & 3.313E-2 & 5.910E-3 \\
			\midrule
			CNN & ConvTimeNet & \textit{WWW 2025} & 1.687 & 4.825E-2 & 1.237E-2 & 1.035E-1 & 4.376E-2 & 7.096E-3 & 1.143E-4 & 3.148E-2 & 3.845E-3 & 1.572E-2 & 1.073E-3 \\
			\midrule
			\multirow{2}{*}{RNN} & LSTM & \textit{EACFM 2022} & \underline{1.377} & 2.328E-2 & 1.138E-3 & 5.201E-2 & 6.087E-3 & 1.078E-2 & 2.865E-4 & 4.936E-2 & 6.782E-3 & 3.447E-2 & 6.320E-3 \\
			& xLSTM & \textit{ICML 2024} & 1.592 & 5.855E-2 & 5.362E-3 & 7.680E-2 & 1.228E-2 & 1.686E-2 & 8.068E-4 & 4.891E-2 & 4.988E-3 & 2.932E-2 & 4.723E-3 \\
			\midrule
			\multirow{2}{*}{Mamba} & Mamba & \textit{COLM 2024} & 1.559 & 2.418E-1 & 8.817E-2 & 5.790E-2 & 7.272E-3 & 2.245E-2 & 2.210E-3 & 1.147E-1 & 2.869E-2 & 5.003E-2 & 1.098E-2 \\
			& Mamba2 & \textit{ICML 2024} & 1.585 & 2.001E-2 & 8.277E-4 & 5.255E-2 & 6.267E-3 & 6.395E-3 & 9.046E-5 & 3.030E-2 & 2.469E-3 & 2.191E-2 & 2.824E-3 \\
			\midrule
			\multirow{4}{*}{\cellcolor{white}LLM} & GPT4TS & \textit{NeurIPS 2023} & 3.457 & 3.415E-2 & 2.792E-3 & 7.528E-2 & 1.246E-2 & 8.287E-3 & 1.757E-4 & 1.033E-1 & 3.332E-2 & 5.359E-2 & 1.119E-2 \\
			& Time-LLM & \textit{ICLR 2024} & 2.708 & 2.302E-2 & 1.058E-3 & 5.733E-2 & 7.420E-3 & 6.617E-3 & 1.034E-4 & 8.354E-2 & 2.140E-2 & 2.244E-2 & 2.261E-3 \\
			& \g{\textbf{LLM4Fluid}} & \g{\textbf{\textit{Ours}}} & \g{\textbf{1.025}} & \g{\textbf{1.849E-2}} & \g{\underline{6.742E-4}} & \g{\underline{5.150E-2}} & \g{\underline{5.946E-3}} & \g{\underline{5.859E-3}} & \g{\underline{7.527E-5}} & \g{\underline{2.656E-2}} & \g{\underline{1.591E-3}} & \g{\underline{1.496E-2}} & \g{\underline{9.118E-4}} \\
			& \g{\textbf{LLM4Fluid + LoRA}} & \g{\textbf{\textit{Ours}}} & \g{5.025} & \g{\underline{1.850E-2}} & \g{\textbf{6.738E-4}} & \g{\textbf{5.138E-2}} & \g{\textbf{5.821E-3}} & \g{\textbf{5.818E-3}} & \g{\textbf{7.306E-5}} & \g{\textbf{2.637E-2}} & \g{\textbf{1.558E-3}} & \g{\textbf{1.374E-2}} & \g{\textbf{8.030E-4}} \\
			\bottomrule
		\end{tabular}
	}

\end{table*}

\begin{table*}[t]
	\centering
	\caption{Zero-shot prediction performance of different models. ``A $\rightarrow$ B'' trains models on dataset A and evaluates on dataset B without retraining.}
	\label{tab:3}
	\scriptsize
	\vskip -0.1cm
	\resizebox{1.0\linewidth}{!}{
	\begin{tabular}{c|c|
			ccccccccc}
		\toprule
		Type & Metric &
		\textbf{LLM4Fluid} & TimeXer & TimeKAN & FilterTS & ConvTimeNet & Time-LLM & GPT4TS & Mamba2 & xLSTM \\
		\midrule
		\multirow{2}{*}{Low-Re $\rightarrow$ High-Re}
		& MAE $\downarrow$ & \g{\textbf{5.229E-2}} & \underline{5.397E-2} & 5.715E-2 & 8.484E-2 & 1.003E-1 & 6.302E-2 & 7.961E-2 & 2.158E-1 & 2.155E-1 \\
		& MSE $\downarrow$ & \g{\textbf{5.921E-3}} & \underline{6.270E-3} & 7.371E-3 & 1.805E-2 & 4.729E-2 & 8.448E-3 & 1.412E-2 & 7.746E-2 & 7.845E-2 \\
		\midrule
		\multirow{2}{*}{High-Re $\rightarrow$ Cavity}
		& MAE $\downarrow$ & \g{\textbf{7.909E-3}} & 8.369E-3 & \underline{8.248E-3} & 5.893E-2 & 5.408E-2 & 8.252E-3 & 1.321E-2 & 1.228E-1 & 1.231E-1 \\
		& MSE $\downarrow$ & \g{\textbf{1.445E-4}} & 1.616E-4 & 1.580E-4 & 2.853E-2 & 6.557E-3 & \underline{1.579E-4} & 5.820E-4 & 3.553E-2 & 3.533E-2 \\
		\midrule
		\multirow{2}{*}{Cavity $\rightarrow$ Channel}
		& MAE $\downarrow$ & \g{\textbf{4.235E-2}} & 6.846E-2 & \underline{4.624E-2} & 6.465E-2 & 5.824E-2 & 4.889E-2 & 5.615E-2 & 1.354E-1 & 1.287E-1 \\
		& MSE $\downarrow$ & \g{\textbf{5.229E-3}} & 1.711E-2 & \underline{6.070E-3} & 1.217E-2 & 1.092E-2 & 6.526E-3 & 1.001E-2 & 3.457E-2 & 3.295E-2 \\
		\midrule
		\multirow{2}{*}{Channel $\rightarrow$ Dam}
		& MAE $\downarrow$ & \g{\textbf{2.561E-2}} & 7.005E-2 & \underline{3.044E-2} & 3.335E-2 & 1.267E-1 & 3.271E-2 & 3.473E-2 & 6.997E-2 & 6.934E-2 \\
		& MSE $\downarrow$ & \g{\textbf{3.770E-3}} & 3.141E-2 & 5.153E-3 & 5.953E-3 & 2.566E-1 & \underline{4.569E-3} & 5.069E-3 & 1.578E-2 & 1.539E-2 \\
		\midrule
		\multirow{2}{*}{Dam $\rightarrow$ Low-Re}
		& MAE $\downarrow$ & \g{\textbf{2.298E-2}} & 4.971E-2 & \underline{2.350E-2} & 3.135E-2 & 3.481E-2 & 2.353E-2 & 5.683E-2 & 1.685E-1 & 1.557E-1 \\
		& MSE $\downarrow$ & \g{\textbf{1.107E-3}} & 8.627E-3 & \underline{1.160E-3} & 3.083E-3 & 5.742E-3 & 1.164E-3 & 9.595E-3 & 4.410E-2 & 3.693E-2 \\
		\bottomrule
	\end{tabular}
	}
	\vskip -0.2cm
\end{table*}

\subsection{Comparative Analysis with SOTA} \label{sec:comparative analysis with sotas}
\noindent\textbf{Reconstruction.}
As shown in \cref{tab:1}, the best reconstruction results are achieved with moderate disentanglement at $\lambda$=1E-4 or 1E-5. We further visualize the correlations among the learned latent representations under different disentanglement weights in \cref{fig:5}. When $\lambda=0$, latent dimensions exhibit strong interdependence, indicating an entangled and redundant latent space. As $\lambda$ increases, cross-dimension correlations gradually decrease, yielding a more orthogonal structure. This suggests that the model separates distinct factors of flow variation and provides a near-orthogonal and physics-disentangled latent space for downstream temporal prediction. In this study, we choose $\lambda$=1E-4, which achieves the best reconstruction quality while maintaining strong orthogonality. Additional latent-space analysis is provided in \cref{sec:latent space analysis}.

\noindent\textbf{Flow field prediction.}
We compare LLM4Fluid with SOTA baselines on five datasets to assess prediction performance. As shown in \cref{tab:2}, LLM4Fluid achieves the best overall performance while using the fewest trainable parameters. On the Dam dataset, LLM4Fluid demonstrates substantial improvements over Time-LLM: MAE decreases by \textbf{33.33\%} and MSE by \textbf{59.67\%}, while the number of trainable parameters is reduced by \textbf{62.15\%}. These results indicate that LLM4Fluid better captures the underlying spatio-temporal evolution of fluid dynamics while maintaining an efficient model design. More detailed visualization results are provided in \cref{sec:flow field prediction}.

\begin{figure}[t]
	\centering
	\includegraphics[width=0.9\linewidth]{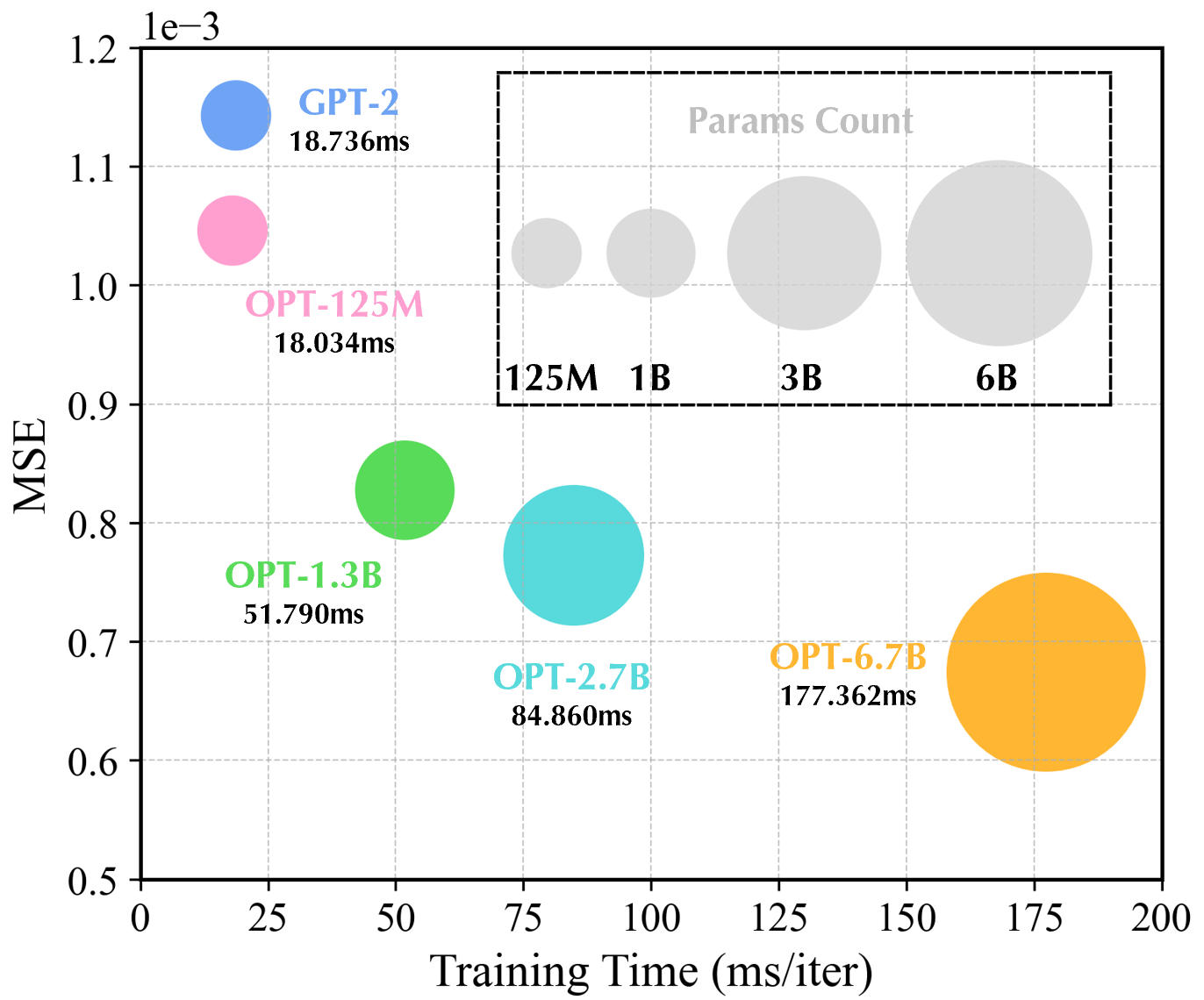}
	\vskip -0.1cm
	\caption{Comparison of accuracy and efficiency for alternative LLM backbones on the Low-Re dataset. LLM4Fluid shows excellent generality and scalability across different LLM backbones.}
	\vskip -0.2cm
	\label{fig:7}
\end{figure}

\begin{figure}[t]
	\centering
	\includegraphics[width=1.0\linewidth]{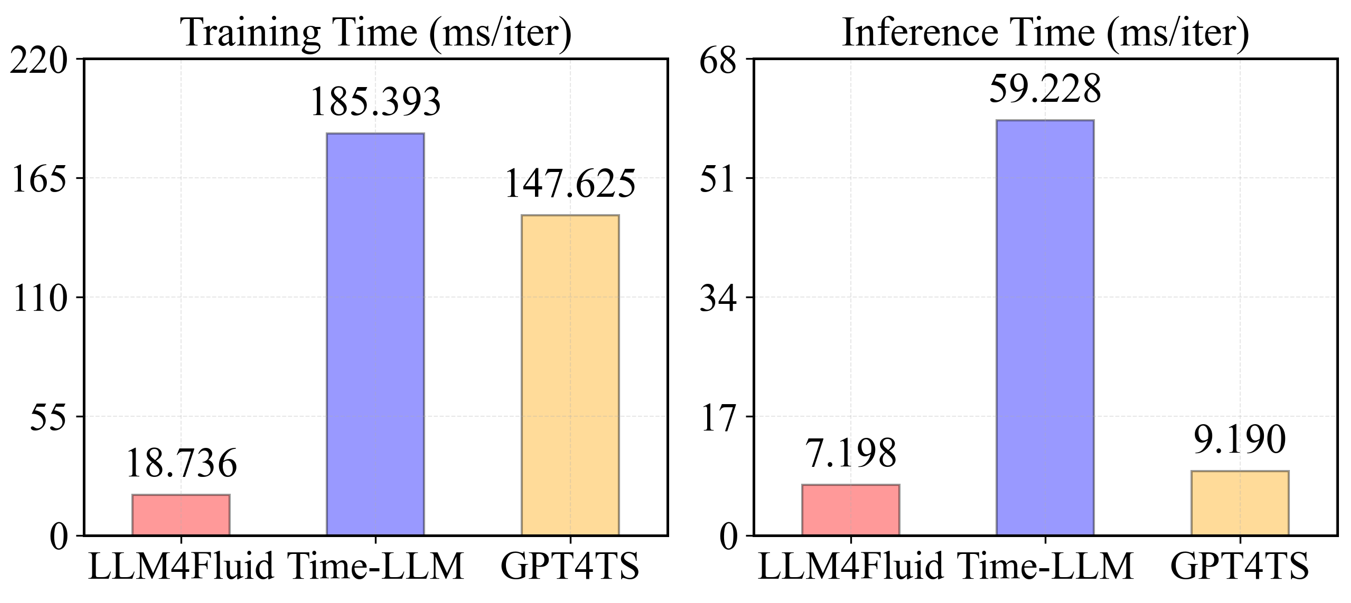}
	\vskip -0.1cm
	\caption{Comparison of computational cost for LLM4Fluid and competing LLM-based methods on the Low-Re dataset using the GPT-2 backbone. LLM4Fluid achieves the lowest cost and shows an obvious efficiency advantage.}
	\vskip -0.2cm
	\label{fig:8}
\end{figure}

\noindent\textbf{Zero-shot prediction.}
We evaluate the zero-shot generalization capability of LLM4Fluid by training on one scenario and evaluating on another without retraining, as illustrated in \cref{fig:3}. As reported in \cref{tab:3}, LLM4Fluid achieves the best performance across all transfer directions. When transferring from Low-Re to High-Re, LLM4Fluid reduces MAE by \textbf{34.32\%} and MSE by \textbf{58.07\%} compared with GPT4TS. These results indicate that LLM4Fluid effectively exploits pretrained LLMs for cross-scenario transfer, outperforming existing methods and generalizing to unseen flows without retraining.

\noindent\textbf{In-context prediction.}
To enhance cross-scenario adaptation, we incorporate physical sequence prompts from the target scenario during prediction. As shown in \cref{fig:6}, in-context prompting consistently reduces SMAPE compared with the zero-shot setting, and performance further improves as more prompts are provided. This demonstrates LLM4Fluid’s in-context learning capability, enabling it to extract transferable cues about the underlying fluid dynamics from the demonstrations.

\noindent\textbf{Generality and scalability.}
LLM4Fluid is compatible with a broad range of LLM backbones, enabling flexible substitution and seamless integration with LLMs of varying capacity. As shown in \cref{fig:7}, LLM4Fluid performs well with GPT-2 and the OPT family at different parameter scales. Moreover, prediction accuracy consistently improves as the backbone size increases, demonstrating that LLM4Fluid can exploit larger models to achieve higher-fidelity predictions. These results confirm the framework’s architectural generality and predictable scaling behavior.

\noindent\textbf{Computational cost.}
As shown in \cref{fig:8}, we compare the training and inference cost of LLM4Fluid against competing LLM-based baselines using the same GPT-2 backbone. Relative to Time-LLM, LLM4Fluid is \textbf{9.9}$\bm{\times}$ faster in training and \textbf{8.2}$\bm{\times}$ faster in inference. Compared with GPT4TS, LLM4Fluid also achieves lower computational cost while using fewer trainable parameters. These results show that lightweight projection layers and a minimized set of trainable components yield substantial efficiency gains without sacrificing prediction accuracy.

\noindent\textbf{LoRA adaptation.}
To further improve adaptation to the target flow scenario, we fine-tune the LLM backbone with LoRA. As shown in \cref{tab:2}, LoRA adds only a small number of trainable parameters yet yields consistent performance gains across all datasets, indicating that limited updates are sufficient to capture scenario-specific dynamics. However, this adaptation involves a trade-off: overly aggressive fine-tuning can reduce the benefits of pretrained priors and increase computational cost. We thus adopt a moderate LoRA configuration to balance adaptation efficiency and generalization. More details are provided in \cref{sec:implementation details}.

\subsection{Ablation Study}\label{ablation study}
We perform ablation experiments on the Low-Re and High-Re datasets to assess the contributions of key components in LLM4Fluid, as shown in \cref{tab:4}. Removing the LLM backbone (w/o LLM) leads to a drastic increase in prediction errors, highlighting the importance of the pretrained LLM for fluid dynamics modeling. Removing prompt-derived positional embeddings (w/o PE) also degrades performance, indicating that the modality alignment strategy is critical for bridging the modality gap and the semantic prompts are essential for providing contextual priors. Replacing the LLM with a single multi-head attention layer (LLM2Attn) or a single Transformer encoder layer (LLM2Trsf) further reduces accuracy, suggesting that shallow attention modules are insufficient to match the representational capacity of the pretrained LLMs. Overall, these results confirm that both the pretrained LLM and prompt-based positional embeddings are crucial, and that the full LLM architecture substantially outperforms lightweight attention alternatives.

\begin{table}[t]
	\centering
	\caption{
		Ablation study of LLM4Fluid. (1) \textit{w/o LLM} removes the LLM backbone and directly feeds tokens into the output projection layer; (2) \textit{w/o PE} removes the position embeddings derived from textual prompts; (3) \textit{LLM2Attn} replaces the LLM with a single multi-head attention layer; (4) \textit{LLM2Trsf} replaces the LLM with a single transformer encoder layer.
	}
	\label{tab:4}
	\scriptsize
	\setlength{\tabcolsep}{3pt}
	\vskip -0.1cm
	\resizebox{1.0\linewidth}{!}{
	\begin{tabular}{@{\extracolsep{\fill}}c|
			ccc|
			ccc}
		\toprule
		\multicolumn{1}{c}{\multirow{2}{*}{Model}} &
		\multicolumn{3}{c}{Low-Re} &
		\multicolumn{3}{c}{High-Re} \\
		\cmidrule(lr){2-4} \cmidrule(lr){5-7}
		\multicolumn{1}{c}{~} & \multicolumn{1}{c}{MAE $\downarrow$} & \multicolumn{1}{c}{MSE $\downarrow$} & \multicolumn{1}{c}{SMAPE $\downarrow$}
		& \multicolumn{1}{c}{MAE $\downarrow$} & \multicolumn{1}{c}{MSE $\downarrow$} & \multicolumn{1}{c}{SMAPE $\downarrow$} \\
		\midrule
		\g{\textbf{LLM4Fluid}}
		& \g{\textbf{1.849E-2}} & \g{\textbf{6.742E-4}} & \g{\textbf{1.829E-1}}
		& \g{\textbf{5.150E-2}} & \g{\textbf{5.946E-3}} & \g{\textbf{4.683E-1}} \\
		
		w/o LLM
		& 1.118E-1 & 1.406E-1 & 4.379E-1
		& 4.273E-1 & 6.894E+0 & 8.279E-1 \\
		
		w/o PE
		& \underline{2.288E-2} & \underline{1.051E-3} & 2.171E-1
		& 5.831E-2 & \underline{7.266E-3} & 5.178E-1 \\
		
		LLM2Attn
		& 2.350E-2 & 1.161E-3 & \underline{2.163E-1}
		& \underline{5.721E-2} & 7.385E-3 & \underline{5.002E-1} \\
		
		LLM2Trsf
		& 2.353E-2 & 1.164E-3 & 2.165E-1
		& 5.739E-2 & 7.437E-3 & 5.008E-1 \\
		\bottomrule
	\end{tabular}
	}
	\vskip -0.2cm
\end{table}

\section{Conclusion}
\label{sec:conclusion}

In this paper, we propose LLM4Fluid, a spatio-temporal prediction framework for fluid dynamics. LLM4Fluid first performs disentangled reduced-order modeling enhanced with a physics-informed disentanglement mechanism, which mitigates spatial feature entanglement while preserving essential flow structures. It then employs a pretrained LLM-based temporal processor with a modality alignment strategy to bridge the semantic–physical modality gap and enable stable long-term prediction. Extensive experiments across diverse flow scenarios demonstrate that LLM4Fluid consistently outperforms SOTA methods, serving as an accurate, efficient, robust, and generalizable neural solver.

Despite these advantages, LLM4Fluid is currently limited to 2D flow datasets with a fixed spatial resolution. Future work will extend it to 3D and multi-physics systems while integrating governing equations into the latent dynamics, further advancing intelligent fluid modeling.


\section*{Impact Statement}
This paper presents work whose goal is to advance the field of machine learning. There are many potential societal consequences of our work, none of which we feel must be specifically highlighted here.

\section*{Acknowledgment}
This research was partially supported by the National Natural Science Foundation of China (12402349), the Natural Science Foundation of Hunan Province (2024JJ6468), the Innovation Reserch Foundation of National University of Defense Technology (ZK2023-11), and the National Key Research and Development Program of China (2021YFB0300101).

\bibliography{example_paper}
\bibliographystyle{icml2026}

\newpage
\appendix

\section{Dataset Details} \label{sec:dataset details}
In this section, we provide the numerical simulation details for all datasets.

\noindent\textbf{Kolmogorov datasets.}
The incompressible Navier-Stokes equations in vorticity form for the Kolmogorov flow are given by:
\begin{equation}
	\frac{\partial \omega(\boldsymbol{x}, t)}{\partial t} + \boldsymbol{u}(\boldsymbol{x}, t) \cdot \nabla \omega(\boldsymbol{x}, t)
	= \frac{1}{Re} \nabla^2 \omega(\boldsymbol{x}, t) + f(\boldsymbol{x}),
	\label{eq:11}
\end{equation}
\begin{equation}
	\nabla \cdot \boldsymbol{u}(\boldsymbol{x}, t) = 0,
	\label{eq:12}
\end{equation}
\begin{equation}
	\omega(\boldsymbol{x}, 0) = \omega_0(\boldsymbol{x}),
	\label{eq:13}
\end{equation}
where $\omega$ denotes the vorticity and $\boldsymbol{x}=(x_1, x_2)  \in(0,2 \pi)^2$ denotes the spatial coordinate. The forcing term is defined as:
\begin{equation}
	f(\boldsymbol{x}) = -k \cos(k x_2) - 0.1 \omega(\boldsymbol{x}),
	\label{eq:14}
\end{equation}
where $k$=8. The simulation is conducted with periodic boundary conditions in all spatial directions. The initial vorticity $\omega_0$ is sampled from a prescribed Gaussian random field.

We modify the pseudo-spectral solver (under MIT License) from 
\href{https://github.com/BaratiLab/FactFormer}{https://github.com/BaratiLab/FactFormer} to generate the data \cite{li2023scalable}. The numerical simulation is performed on a uniform grid of 2048$\times$2048 with a temporal resolution of 1E-4. Snapshots are recorded with $\Delta t=0.001$ s, and downsampled to a spatial resolution of 128$\times$128. Each dataset consists of 1500 snapshots for one trajectory.

\noindent\textbf{Cavity dataset.}
The lid-driven cavity flow is governed by the incompressible Navier--Stokes equations:
\begin{equation}
	\frac{\partial \boldsymbol{u}}{\partial t} + \boldsymbol{u}\cdot\nabla \boldsymbol{u}
	= -\nabla p + \nu \nabla^2 \boldsymbol{u},
\end{equation}
\begin{equation}
	\nabla \cdot \boldsymbol{u} = 0,
\end{equation}
where $\boldsymbol{u}=(u,v,w)$ is the velocity, $p$ is the kinematic pressure, and $\nu=0.0001\ \mathrm{m^2/s}$ is the kinematic viscosity.
We generate the data using the standard OpenFOAM-10 tutorial
\texttt{cavity} with the transient solver \texttt{icoFoam},
which advances the incompressible equations using a pressure--velocity coupling scheme.
The domain is $x,y\in[0,0.1]$ m and $z\in[0,0.01]$ m, discretized by a $128\times128\times1$ mesh. The front and back boundaries are set to empty, and snapshots are extracted from the slice $z=0.005$ m,
yielding 2D fields on the $x$--$y$ plane. We keep other settings consistent with the original tutorial case.
The lid velocity is set to 1 m/s. The simulation runs over $t\in[0,1.5]$ s with a time step of $\Delta t=0.001$ s, and we record 1500 snapshots for one trajectory.

\noindent\textbf{Channel dataset.}
The turbulent channel flow is modeled by the filtered incompressible Navier--Stokes equations for large eddy simulation (LES):
\begin{equation}
	\frac{\partial \bar{\boldsymbol{u}}}{\partial t} + \bar{\boldsymbol{u}}\cdot\nabla \bar{\boldsymbol{u}}
	= -\nabla \bar{p} + \nu \nabla^2 \bar{\boldsymbol{u}} - \nabla\cdot \boldsymbol{\tau}_{\mathrm{sgs}},
\end{equation}
\begin{equation}
	\nabla\cdot \bar{\boldsymbol{u}} = 0,
\end{equation}
where $\bar{\boldsymbol{u}}$ and $\bar{p}$ denote the filtered velocity and pressure, and $\boldsymbol{\tau}_{\mathrm{sgs}}$ is the sub-grid scale stress.
We use the OpenFOAM-10 tutorial
\texttt{channel395} and compute the flow with the transient solver \texttt{pimpleFoam},
which employs the PIMPLE algorithm (a hybrid PISO--SIMPLE procedure) for pressure--velocity coupling and supports LES modeling. The bulk velocity is set to $\bar{U}=0.1335\ \mathrm{m/s}$ and the viscosity is $\nu=2\times 10^{-5}\ \mathrm{m^2/s}$. The 3D computational domain is $x\in[0,4]$ m and $y,z\in[0,2]$ m with a mesh of $80\times50\times60$, with grid refinement near the top and bottom walls. The simulation runs over $t\in[0,15]$ s with a time step of $\Delta t=0.01$ s, and we record 1500 snapshots. For modeling, we extract a 2D slice at $x=2$ m (the $y$--$z$ plane) and interpolate the fields to $128\times128$. Other settings follow the original tutorial case.

\noindent\textbf{Dam dataset.}
The dam-break flow is a two-phase incompressible system (water--air) modeled by the volume-of-fluid (VOF) formulation:
\begin{equation}
	\nabla\cdot \boldsymbol{u} = 0,
\end{equation}
\begin{equation}
	\frac{\partial (\rho \boldsymbol{u})}{\partial t} + \nabla\cdot(\rho \boldsymbol{u}\otimes \boldsymbol{u})
	= -\nabla p + \nabla\cdot\big(\mu(\nabla \boldsymbol{u} + \nabla \boldsymbol{u}^{\top})\big)
	+ \rho \boldsymbol{g} + \boldsymbol{f}_{\sigma},
\end{equation}
\begin{equation}
	\frac{\partial \alpha}{\partial t} + \nabla\cdot(\alpha \boldsymbol{u}) = 0,
\end{equation}
where $\alpha\in[0,1]$ is the volume fraction of water, and the mixture properties are
$\rho = \alpha \rho_w + (1-\alpha)\rho_a$ and $\mu = \alpha \mu_w + (1-\alpha)\mu_a$.
The gravity is $\boldsymbol{g}=(0,-9.81,0)\ \mathrm{m/s^2}$ and $\boldsymbol{f}_{\sigma}$ denotes the surface-tension force with $\sigma=0.07$.
We generate the data using the OpenFOAM-10 tutorial
\texttt{damBreak} with the VOF solver \texttt{interFoam},
which solves the phase-fraction advection and the incompressible momentum equations with surface tension.
We set the initial velocity to zero. The air properties are $\nu_a=1.48\times 10^{-5}\ \mathrm{m^2/s}$ and $\rho_a=1\ \mathrm{kg/m^3}$, and the water properties are $\nu_w=10^{-6}\ \mathrm{m^2/s}$ and $\rho_w=10^3\ \mathrm{kg/m^3}$. The domain is $x,y\in[0,0.584]$ m and $z\in[0,0.0146]$ m. Compared with the original case, we double the base mesh resolution and apply local refinement in the upper and lateral regions around the dam-break, while keeping other settings unchanged. The simulation runs over $t\in[0.1,0.25]$ s with a time step of $\Delta t=0.0001$ s, and we record 1500 snapshots. We extract 2D fields from the slice $z=0.0073$ m (the $x$--$y$ plane) and interpolate them to $128\times128$ for learning.

\begin{algorithm*}[th!]
	\caption{Training stages of LLM4Fluid}
	\label{alg:1}
	\begin{algorithmic}[1]
		\STATE \textbf{Input:} Training set $\mathbf{X}_{1:T}$, 
		Encoder $\mathcal{E}_\theta$, Decoder $\mathcal{D}_\theta$, Frozen LLM backbone $\mathcal{F}_\phi$, 
		Batch size $B$, Sliding window length $M$, Patch length $M_p$, Number of patches $N = \left\lfloor \frac{M}{M_p} \right\rfloor$, Latent space dimension $D$, Embedding dimension $D_e$, Number of epochs $E$, Training iterations $S = \left\lceil \frac{T}{B} \right\rceil$, Optimizers $\text{Opt}_\theta$ and $\text{Opt}_\psi$
		\STATE \textbf{Output:} Trained reduced-order model parameters $\theta^\star$ and temporal processor parameters $\psi^\star$
		\vspace{2pt}
		\STATE \textbf{// Stage 1: Disentangled reduced-order modeling}
		\FOR{epoch = 1 to $E$}
		\FOR{iteration = 1 to $S$}
		\STATE Initialize batch loss: $\mathcal{L} \leftarrow 0$
		\FOR{$b$ = 1 to $B$}
		\STATE Encode flow field snapshot: $(\boldsymbol{\mu}_b, \boldsymbol{\sigma}_b) \leftarrow \mathcal{E}_\theta(\mathbf{X}_b)$  \Comment{$\mathbf{X}_b \in \mathbb{R}^{H \times W \times C}$}
		\STATE Sample latent variables: $\mathbf{z}_b \leftarrow \boldsymbol{\mu}_b + \boldsymbol{\sigma}_b \odot \boldsymbol{\epsilon}_b$, with $\boldsymbol{\epsilon}_b \sim \mathcal{N}(\mathbf{0}, \mathbf{I})$  \Comment{$\mathbf{z}_b, \boldsymbol{\mu}_b, \boldsymbol{\sigma}_b, \boldsymbol{\epsilon}_b \in \mathbb{R}^{D}$}
		\STATE Reconstruct snapshot: $\widehat{\mathbf{X}}_b \leftarrow \mathcal{D}_\theta(\mathbf{z}_b)$  \Comment{$\widehat{\mathbf{X}}_b \in \mathbb{R}^{H \times W \times C}$}
		\STATE Compute loss: $\mathcal{L}_b \leftarrow \mathcal{L}_{\text{rec}}(\widehat{\mathbf{X}}_b, \mathbf{X}_b) + \lambda\, \mathcal{L}_{\text{dis}}(\boldsymbol{\mu}_b,\boldsymbol{\sigma}_b)$ using \cref{eq:3}.
		\STATE Accumulate loss: $\mathcal{L} \leftarrow \mathcal{L} + \mathcal{L}_b$
		\ENDFOR
		\STATE Compute average batch loss: $\mathcal{L}_{avg} = \mathcal{L}/B$
		\STATE Update reduced-order model parameters: $\theta \leftarrow \text{Opt}_\theta(\theta, \nabla_\theta \mathcal{L}_{avg})$
		\ENDFOR
		\ENDFOR
		\STATE Freeze $\mathcal{E}_\theta$ and $\mathcal{D}_\theta$ with $\theta^\star$, Freeze $\mathcal{F}_\phi$ with $\phi$
		
		\vspace{2pt}
		\STATE \textbf{// Stage 2: LLM-based temporal processor}
		\FOR{epoch = 1 to $E$}
		\FOR{iteration = 1 to $S$}
		\STATE Initialize batch loss: $\mathcal{L} \leftarrow 0$
		\FOR{$b$ = 1 to $B$}
		\STATE Encode flow field snapshot: $(\boldsymbol{\mu}_b, \boldsymbol{\sigma}_b) \leftarrow \mathcal{E}_{\theta^\star}(\mathbf{X}_b)$, set $\mathbf{z}_b \leftarrow \boldsymbol{\mu}_b$
		\STATE Construct physical sequence $\mathbf{z}$ within a window of length $M$  \Comment{$\mathbf{z} \in \mathbb{R}^{D \times M}$}
		\FOR{$i$ = 1 to $D$}
		\STATE Apply RevIN to obtain the normalized $i$-th latent variable $\mathbf{z}^{(i)}$  \Comment{$\mathbf{z}^{(i)} \in \mathbb{R}^{M}$}
		\STATE Divide $\mathbf{z}^{(i)}$ into patches to produce physical token sequence $\mathbf{z}^{(i)}_{1:N}$  \Comment{$\mathbf{z}^{(i)}_{1:N} \in \mathbb{R}^{M_p \times N}$}
		\STATE Project tokens into physical embeddings: $\mathbf{s}^{(i)}_{1:N}=\mathrm{InputProjection}_\psi(\mathbf{z}^{(i)}_{1:N})$  \Comment{$\mathbf{s}^{(i)}_{1:N} \in \mathbb{R}^{D_e \times N}$}
		\STATE Obtain position embeddings from prompts: $\mathbf{k}_{1:N}=\mathrm{Last}(\mathcal{F}_\phi(\mathrm{Tokenizer}(\mathrm{Text}_{1:N})))$  \Comment{$\mathbf{k}_{1:N} \in \mathbb{R}^{D_e \times N}$}
		\STATE Perform modality alignment: $\mathbf{e}^{(i)}_{1:N} \leftarrow \mathbf{s}^{(i)}_{1:N} + \gamma \cdot \mathbf{k}_{1:N}$  \Comment{$\mathbf{e}^{(i)}_{1:N} \in \mathbb{R}^{D_e \times N}$}
		\STATE Feed tokens into frozen LLM: $\widehat{\mathbf{e}}^{(i)}_{N+1:2N} \leftarrow \mathcal{F}_\phi(\mathbf{e}^{(i)}_{1:N})$   \Comment{$\widehat{\mathbf{e}}^{(i)}_{N+1:2N} \in \mathbb{R}^{D_e \times N}$}
		\STATE Project back to latent space: $\widehat{\mathbf{z}}^{(i)}_{N+1:2N} \leftarrow \mathrm{OutputProjection}_\psi(\widehat{\mathbf{e}}^{(i)}_{N+1:2N})$  \Comment{$\widehat{\mathbf{z}}^{(i)}_{N+1:2N} \in \mathbb{R}^{D \times N}$}
		\STATE Compute loss: $\mathcal{L}_{i} \leftarrow \mathcal{L}_{\text{mse}}(\widehat{\mathbf{z}}^{(i)}_{N+1:2N}, \mathbf{z}^{(i)}_{N+1:2N})$ using \cref{eq:9}.
		\STATE Accumulate loss: $\mathcal{L} \leftarrow \mathcal{L} + \mathcal{L}_i$ 
		\ENDFOR
		\ENDFOR
		\STATE Compute average batch loss: $\mathcal{L}_{avg} = \mathcal{L}/B$
		\STATE Update temporal processor parameters: $\psi \leftarrow \text{Opt}_\psi(\psi, \nabla_\psi \mathcal{L}_{avg})$
		\ENDFOR
		\ENDFOR
		\STATE \textbf{Return} $\theta^\star, \psi^\star$
	\end{algorithmic}
\end{algorithm*}

\begin{algorithm*}[th!]
	\caption{Inference stage of LLM4Fluid}
	\label{alg:2}
	\begin{algorithmic}[1]
		\STATE \textbf{Input:} Trained encoder $\mathcal{E}_{\theta^\star}$, decoder $\mathcal{D}_{\theta^\star}$, temporal processor $\mathcal{G}_{\psi^\star}$ with frozen LLM backbone $\mathcal{F}_\phi$, lookback sequence $\mathbf{X}_{1:T}$, sliding window length $M$, prediction horizon $F$
		\STATE \textbf{Output:} Predicted flow fields $\widehat{\mathbf{X}}_{T+1:T+F}$
		\STATE \textbf{// No parameter update during inference}
		\STATE Encode lookback sequence into latent physical sequence: $\mathbf{z}_{1:T} \leftarrow \mathcal{E}_{\theta^\star}(\mathbf{X}_{1:T})$
		\FOR{$t = T+1$ to $T+F$}
		\STATE Construct latent input window $\mathbf{z}_{t-M:t-1}$
		\STATE Predict next physical sequence with LLM-based temporal processor: $\widehat{\mathbf{z}}_{t:t+M-1} \leftarrow \mathcal{G}_{\psi^\star}(\mathbf{z}_{t-M:t-1})$
		\STATE Append $\widehat{\mathbf{z}}_t$ to sequence and update the sliding window for autoregressive prediction
		\ENDFOR
		\STATE Decode predicted flow field snapshots: $\widehat{\mathbf{X}}_{T+1:T+F} \leftarrow \mathcal{D}_{\theta^\star}(\widehat{\mathbf{z}}_{T+1:T+F})$
		\STATE \textbf{Return} $\widehat{\mathbf{X}}_{T+1:T+F}$
	\end{algorithmic}
\end{algorithm*}

\section{Evaluation Metrics} \label{sec:evaluation metrics}
To evaluate the prediction performance of the models on spatio-temporal flow fields, we adopt six complementary metrics. These metrics jointly characterize both accuracy and quality of the predicted flow fields. The definitions are given below.

\begin{itemize}
	\item \textbf{Mean absolute error (MAE).}
	MAE measures the average magnitude of point-wise errors and directly reflects the overall deviation of the predicted flow field from the ground truth:
	\begin{equation}
		\text{MAE} = \frac{1}{N} \sum_{i=1}^N \left| x_i - \hat{x}_i \right|.
		\label{eq15}
	\end{equation}
	
	\item \textbf{Mean squared error (MSE).}
	MSE penalizes larger prediction errors more heavily and is particularly sensitive to outliers, which helps assess whether the model can avoid large local errors in the flow field:
	\begin{equation}
		\text{MSE} = \frac{1}{N} \sum_{i=1}^N \left( x_i - \hat{x}_i \right)^2.
		\label{eq16}
	\end{equation}
	
	\item \textbf{Symmetric mean absolute percentage error (SMAPE).}
	SMAPE evaluates the relative error between predictions and ground truth and is robust to varying magnitudes across different flow regions, making it suitable for flows with both high-intensity and low-intensity structures:
	\begin{equation}
		\text{SMAPE} = \frac{100\%}{N} \sum_{i=1}^N
		\frac{\left| x_i - \hat{x}_i \right|}
		{\left( \left|x_i\right| + \left|\hat{x}_i\right| \right)/2 }.
		\label{eq17}
	\end{equation}
	
	\item \textbf{Coefficient of determination ($\text{R}^\text{2}$).}
	$\text{R}^\text{2}$ measures how much of the variance in the true flow field is explained by the prediction and thus reflects the global goodness-of-fit of the predicted dynamics:
	\begin{equation}
		\text{R}^\text{2} = 1 -
		\frac{\displaystyle\sum_{i=1}^N \left( x_i - \hat{x}_i \right)^2}
		{\displaystyle\sum_{i=1}^N \left( x_i - \bar{x} \right)^2 }.
		\label{eq18}
	\end{equation}
	
	\item \textbf{Peak signal-to-noise ratio (PSNR).}
	PSNR is an image-quality metric that compares the maximum possible signal power with the reconstruction error, highlighting the fidelity of fine-scale spatial details in the predicted flow fields:
	\begin{equation}
		\text{PSNR} = 10 \cdot \log_{10}
		\frac{\left(2^b - 1\right)^2}{\mathrm{MSE}}.
		\label{eq19}
	\end{equation}
	
	\item \textbf{Structural similarity index measure (SSIM).}
	SSIM assesses the similarity of local structures between predictions and ground truth in terms of luminance, contrast, and structure, which is crucial for capturing coherent flow patterns such as vortices and shear layers:
	\begin{equation}
		\text{SSIM} =
		\frac{\left(2 \mu_x \mu_{\hat{x}} + C_1\right)
			\left(2 \sigma_{x \hat{x}} + C_2\right)}
		{\left(\mu_x^2 + \mu_{\hat{x}}^2 + C_1\right)
			\left(\sigma_x^2 + \sigma_{\hat{x}}^2 + C_2\right)}.
		\label{eq20}
	\end{equation}
\end{itemize}

Here, $N$ denotes the total number of data points, and $x_i$ and $\hat{x}_i$ represent the ground truth and predicted values at the $i$-th point, respectively. For the $\text{R}^\text{2}$ metric, $\bar{x}$ denotes the mean of the ground truth. In the PSNR formula, $b$ denotes the bit depth of the pixel values. For SSIM, $\mu_x$ and $\mu_{\hat{x}}$ represent the means of the ground truth and predicted values, respectively; $\sigma_{x \hat{x}}$ denotes their covariance; $\sigma_x^2$ and $\sigma_{\hat{x}}^2$ denote their variances; and $C_1$ and $C_2$ are constants.

\section{Implementation Details} \label{sec:implementation details}
In this section, we summarize the training and inference procedures of LLM4Fluid and present the implementation details required for reproduction.

\cref{alg:1,alg:2} describe the two-stage training pipeline and the autoregressive inference process of LLM4Fluid. In the first training stage, the encoder--decoder $(\mathcal{E}_\theta,\mathcal{D}_\theta)$ is optimized as a disentangled reduced-order model: high-dimensional flow field snapshots $\mathbf{X}_t$ are compressed into near-orthogonal and physics-disentangled latent variables $\mathbf{z}_t$ by minimizing the reconstruction loss in \cref{eq:2} together with the disentanglement loss in \cref{eq:3}. After training, the parameters $\theta$ are frozen, and $(\mathcal{E}_{\theta^\star},\mathcal{D}_{\theta^\star})$ are used as a reduced-order surrogate that provides stable latent representations and reconstructs flow fields from the latent space. In the second training stage, we train the LLM-based temporal processor $\mathcal{G}_\psi$ on sequences of latent variables. For each batch, the snapshots are first encoded by $\mathcal{E}_{\theta^\star}$, and a sliding window of length $M$ is constructed. Each latent physical sequence is then normalized (RevIN), divided into patches, projected into embeddings, and combined with LLM-embedded prompts to form aligned physical embeddings, which are processed by the frozen LLM backbone $\mathcal{F}_\phi$. Only the lightweight modules surrounding the backbone (including RevIN, input projection, modality alignment, and output projection) are updated, while $\mathcal{F}_\phi$ remains fixed. This stage is trained with the MSE loss between the predicted and ground-truth future latent variables in \cref{eq:9}. During inference, given a lookback flow sequence $\mathbf{X}_{1:T}$, we first encode it with the frozen encoder $\mathcal{E}_{\theta^\star}$ to obtain the latent physical sequence $\mathbf{z}_{1:T}$. Then, as summarized in \cref{alg:2}, we perform an autoregressive rollout: at each step $t$ from $T+1$ to $T+F$, a sliding window of length $M$ is formed from the most recent latent variables and passed through the temporal processor $\mathcal{G}_{\psi^\star}$ to predict the next $M$ latent states $\widehat{\mathbf{z}}_{t:t+M-1}$. Among them,  $\widehat{\mathbf{z}}_{t}$ is appended to the sequence while the oldest latent in the window is discarded. After $F$ steps, all predicted latent variables $\widehat{\mathbf{z}}_{T+1:T+F}$ are decoded by $\mathcal{D}_{\theta^\star}$ to yield the final flow field predictions $\widehat{\mathbf{X}}_{T+1:T+F}$.

In the second training stage, we employ parameter-efficient fine-tuning (PEFT) using Low-Rank Adaptation (LoRA) to adapt the frozen LLM backbone. Specifically, LoRA adapters are injected into the query projection layers (\texttt{q\underline{~}proj}) of each self-attention block within the OPT-6.7B backbone. The LoRA rank is set to $r$=4, the scaling factor to $\alpha$=16, and the LoRA dropout to 0, while the original bias parameters remain frozen (\texttt{bias="none"}). After wrapping the backbone with LoRA, only the LoRA adapter weights and the aforementioned lightweight modules are trainable, whereas all other LLM parameters remain fixed. This configuration enables LLM4Fluid to effectively specialize for flow field prediction with only a few additional parameters and little extra computational cost.

All implementations are built upon the Time Series Library \cite{wang2024timeserieslibrary}, OpenLTM \cite{liu2025openltm}, and other official code repositories of the baselines. To ensure a fair comparison, we keep the original architectural designs and training strategies whenever possible, and only adjust factors such as the number of blocks and hidden dimensions so that all models have a comparable number of parameters. Moreover, all baselines share the same data preprocessing, sliding window strategy, and autoregressive evaluation protocol as LLM4Fluid. Notably, Time-LLM requires a dataset-specific textual prompt to effectively guide its predictions. For the Low-Re dataset, we use the following prompt:``The 2D Kolmogorov turbulence dataset describes the temporal evolution of fluid dynamics at Reynolds number Re = 100, with physical variables including streamwise velocity, vertical velocity, and vorticity, which are encoded by the reduced-order model into 32-dimensional latent vectors for each snapshot." For the High-Re dataset, we use the same prompt but replace ``Re = 100" with ``Re = 1000". For the Cavity, Channel, and Dam datasets, we use prompts with the same suffix while changing only the dataset-specific prefix to
``The 2D cavity flow dataset describes the temporal evolution of incompressible flow dynamics in a lid-driven cavity configuration,''
``The 2D channel flow dataset describes the temporal evolution of incompressible flow dynamics in a channel configuration,''
and ``The 2D dam flow dataset describes the temporal evolution of transient flow dynamics during a dam-break process,'' respectively.

All experiments are conducted on a single NVIDIA RTX 5090 GPU with 32~GB memory. The entire experimental pipeline is implemented in Python~3.12.11 with PyTorch~2.8.0+cu128. Unless otherwise stated, optimization hyperparameters follow the default settings provided in the corresponding repositories.

\begin{table}[t]
	\centering
	\caption{Prediction performance of LLM4Fluid for different prediction horizons on the Low-Re dataset. ``10-10" indicates 10 input steps to predict 10 output steps.}
	\label{tab:5}
	\scriptsize
	\setlength{\tabcolsep}{3pt}
	\resizebox{1.0\linewidth}{!}{
		\begin{tabular*}{\linewidth}{@{\extracolsep{\fill}}c|
				ccc|
				ccc}
			\toprule
			\multicolumn{1}{c}{\multirow{2}{*}{\makecell[c]{Prediction\\horizon}}} &
			\multicolumn{3}{c}{Error} &
			\multicolumn{3}{c}{Quality} \\
			\cmidrule(lr){2-4} \cmidrule(lr){5-7}
			\multicolumn{1}{c}{~} &
			\multicolumn{1}{c}{MAE $\downarrow$} &
			\multicolumn{1}{c}{MSE $\downarrow$} &
			\multicolumn{1}{c}{SMAPE $\downarrow$} &
			\multicolumn{1}{c}{$\text{R}^\text{2}$ $\uparrow$} &
			\multicolumn{1}{c}{PSNR $\uparrow$} &
			\multicolumn{1}{c}{SSIM $\uparrow$} \\
			\midrule
			10--10 & 2.093E-2 & 8.975E-4 & 1.991E-1 & 0.991 & 30.798 & 0.924 \\
			30--30 & \textbf{1.826E-2} & \textbf{6.565E-4} & \textbf{1.812E-1} & \textbf{0.993} & \textbf{31.997} & \textbf{0.936} \\
			20--20 & \underline{1.849E-2} & \underline{6.742E-4} & \underline{1.829E-1} & \underline{0.993} & \underline{31.880} & \underline{0.934} \\
			20--40 & 1.973E-2 & 7.618E-4 & 1.923E-1 & 0.992 & 31.607 & 0.928 \\
			20--60 & 2.163E-2 & 9.262E-4 & 2.077E-1 & 0.991 & 31.073 & 0.917 \\
			20--80 & 2.491E-2 & 1.402E-3 & 2.290E-1 & 0.986 & 30.325 & 0.898 \\
			\bottomrule
		\end{tabular*}
	}
\end{table}

\begin{table}[t]
	\centering
	\caption{Impact of latent dimension $D$ for reconstruction performance on the Low-Re dataset.}
	\label{tab:6}
	\scriptsize
	\setlength{\tabcolsep}{3pt}
	\resizebox{1.0\linewidth}{!}{
		\begin{tabular}{@{\extracolsep{\fill}}c|
				ccc|
				ccc}
			\toprule
			\multicolumn{1}{c}{\multirow{2}{*}{Dimension}} &
			\multicolumn{3}{c}{Error} &
			\multicolumn{3}{c}{Quality} \\
			\cmidrule(lr){2-4} \cmidrule(lr){5-7}
			\multicolumn{1}{c}{~} &
			\multicolumn{1}{c}{MAE $\downarrow$} &
			\multicolumn{1}{c}{MSE $\downarrow$} &
			\multicolumn{1}{c}{SMAPE $\downarrow$} &
			\multicolumn{1}{c}{R2 $\uparrow$} &
			\multicolumn{1}{c}{PSNR $\uparrow$} &
			\multicolumn{1}{c}{SSIM $\uparrow$} \\
			\midrule
			8   & 2.917E-2 & 1.645E-3 & 2.611E-1 & 0.984 & 27.892 & 0.872 \\
			16  & 2.471E-2 & 1.272E-3 & 2.241E-1 & 0.987 & 29.117 & 0.905 \\
			32  & \textbf{1.958E-2} & \textbf{7.735E-4} & \textbf{1.899E-1} & \textbf{0.992} & \textbf{31.351} & \textbf{0.929} \\
			64  & \underline{2.232E-2} & \underline{1.066E-3} & \underline{2.107E-1} & \underline{0.989} & \underline{29.880} & \underline{0.912} \\
			128 & 2.395E-2 & 1.279E-3 & 2.208E-1 & 0.987 & 29.056 & 0.907 \\
			\bottomrule
		\end{tabular}
	}
\end{table}

\begin{table}[t]
	\centering
	\caption{Prediction performance of LLM4Fluid with different LLM backbones on the Low-Re dataset.}
	\label{tab:7}
	\scriptsize
	\setlength{\tabcolsep}{3pt}
	\resizebox{1.0\linewidth}{!}{
		\begin{tabular}{@{\extracolsep{\fill}}c|
				ccc|
				ccc}
			\toprule
			\multicolumn{1}{c}{\multirow{2}{*}{LLM}} &
			\multicolumn{3}{c}{Error} &
			\multicolumn{3}{c}{Quality} \\
			\cmidrule(lr){2-4} \cmidrule(lr){5-7}
			\multicolumn{1}{c}{~} &
			\multicolumn{1}{c}{MAE $\downarrow$} &
			\multicolumn{1}{c}{MSE $\downarrow$} &
			\multicolumn{1}{c}{SMAPE $\downarrow$} &
			\multicolumn{1}{c}{R2 $\uparrow$} &
			\multicolumn{1}{c}{PSNR $\uparrow$} &
			\multicolumn{1}{c}{SSIM $\uparrow$} \\
			\midrule
			GPT-2      & 2.331E-2 & 1.143E-3 & 2.151E-1 & 0.989 & 29.953 & 0.914 \\
			OPT-125M   & 2.248E-2 & 1.046E-3 & 2.092E-1 & 0.990 & 30.266 & 0.918 \\
			OPT-1.3B   & 2.063E-2 & 8.274E-4 & 2.008E-1 & \underline{0.992} & 31.290 & 0.921 \\
			OPT-2.7B   & \underline{1.998E-2} & \underline{7.726E-4} & \underline{1.957E-1} & \underline{0.992} & \underline{31.476} & \underline{0.925} \\
			OPT-6.7B   & \textbf{1.849E-2} & \textbf{6.742E-4} & \textbf{1.829E-1} & \textbf{0.993} & \textbf{31.880} & \textbf{0.934} \\
			\bottomrule
		\end{tabular}
	}
\end{table}

\section{Supplementary Results} \label{sec:supplementary results}

\subsection{Prediction Horizon} \label{sec:prediction horizon}
We evaluate LLM4Fluid under different input–output horizons on the Low-Re dataset, as shown in \cref{tab:5}. Here, ``10–10" denotes using 10 historical steps to predict the next 10 steps (input:output = 1:1), while ``20–80" uses 20 historical steps to predict 80 future steps (input:output = 1:4). For the configurations with a fixed 20-step input (20–20, 20–40, 20–60, 20–80), increasing the prediction horizon leads to gradually higher MAE, MSE and SMAPE and lower $\text{R}^\text{2}$, PSNR and SSIM, reflecting the accumulation of errors over longer rollouts. Notably, the 30–30 setting achieves the best prediction performance across both error and quality metrics, suggesting that richer temporal context can partially offset the increased difficulty of long-horizon prediction.

\begin{figure}[t]
	\centering
	\includegraphics[width=0.9\linewidth]{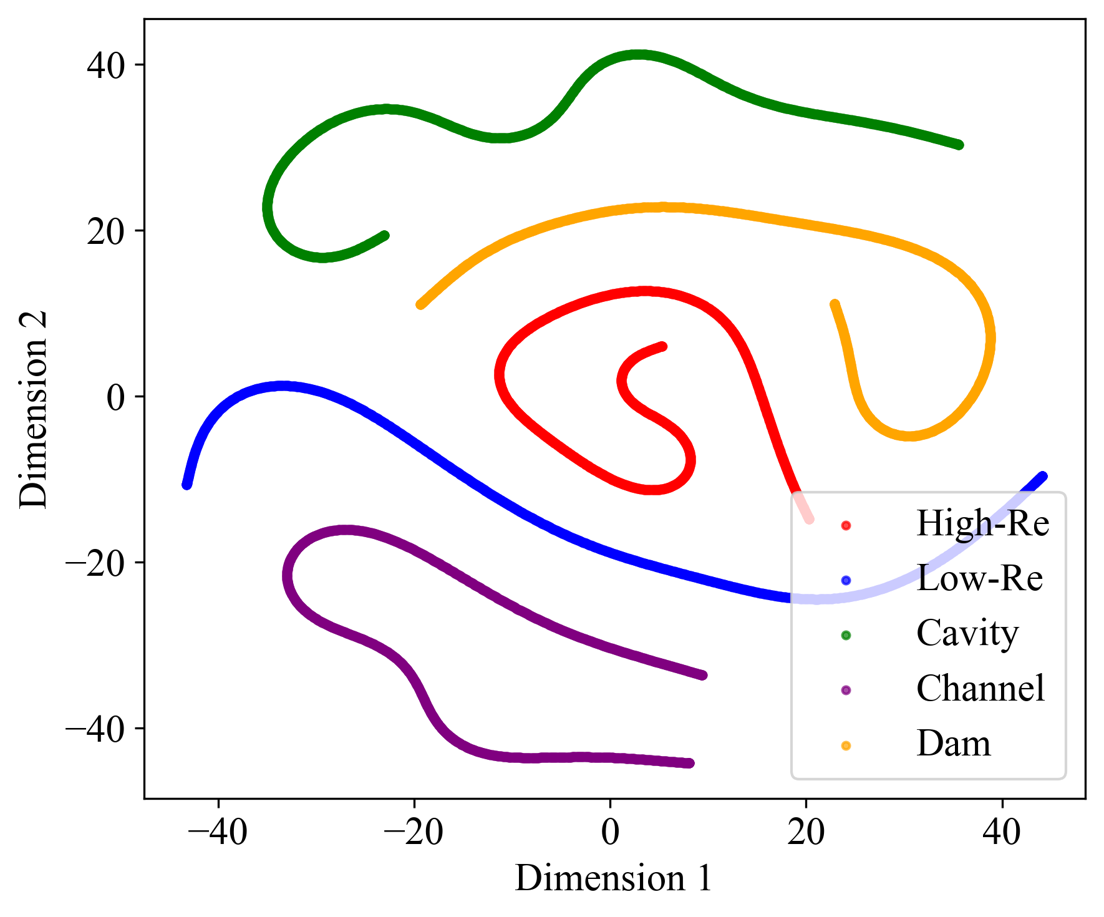}
	\vskip -0.1cm
	\caption{The t-SNE visualization of latent representations on five datasets. The learned latent space exhibits clear separation across different flow scenarios.}
	\label{fig:9}
	\vskip -0.2cm
\end{figure}

\begin{figure*}[t]
	\centering
	\begin{subfigure}[b]{0.49\linewidth}
		\centering
		\includegraphics[width=\linewidth]{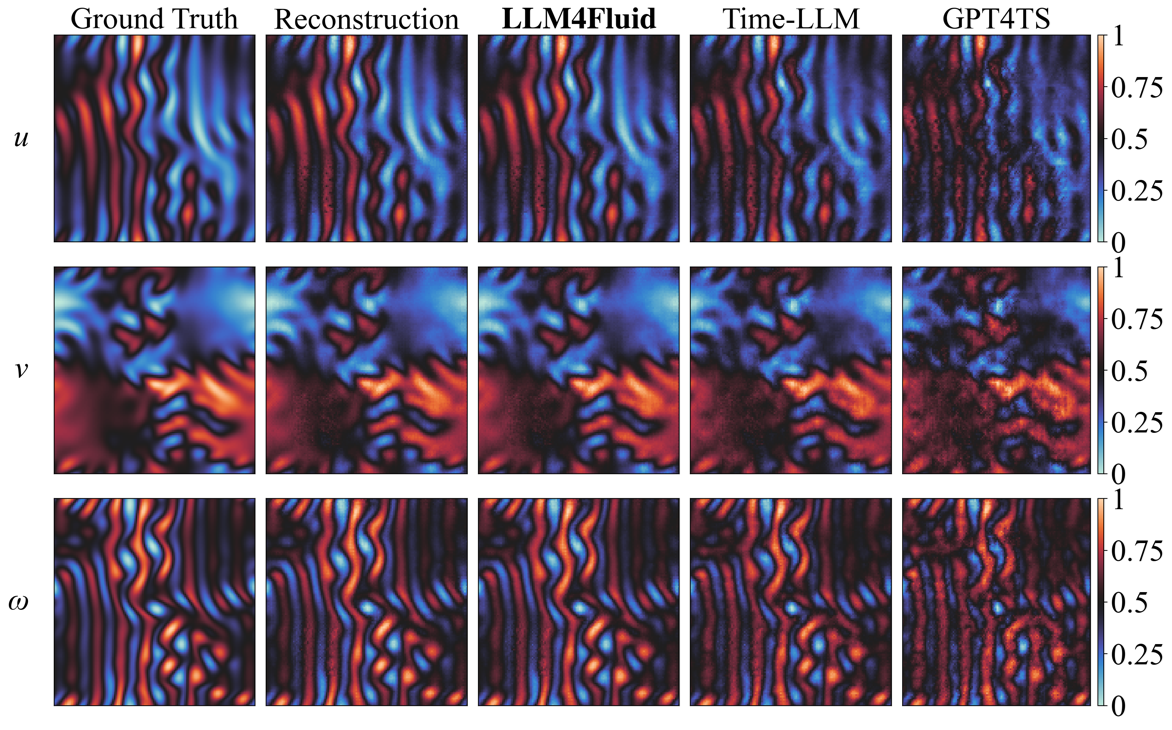}
		\subcaption{}
		\label{fig:10a}
	\end{subfigure}
	\hspace{10pt}%
	\begin{subfigure}[b]{0.427\linewidth}
		\centering
		\includegraphics[width=\linewidth]{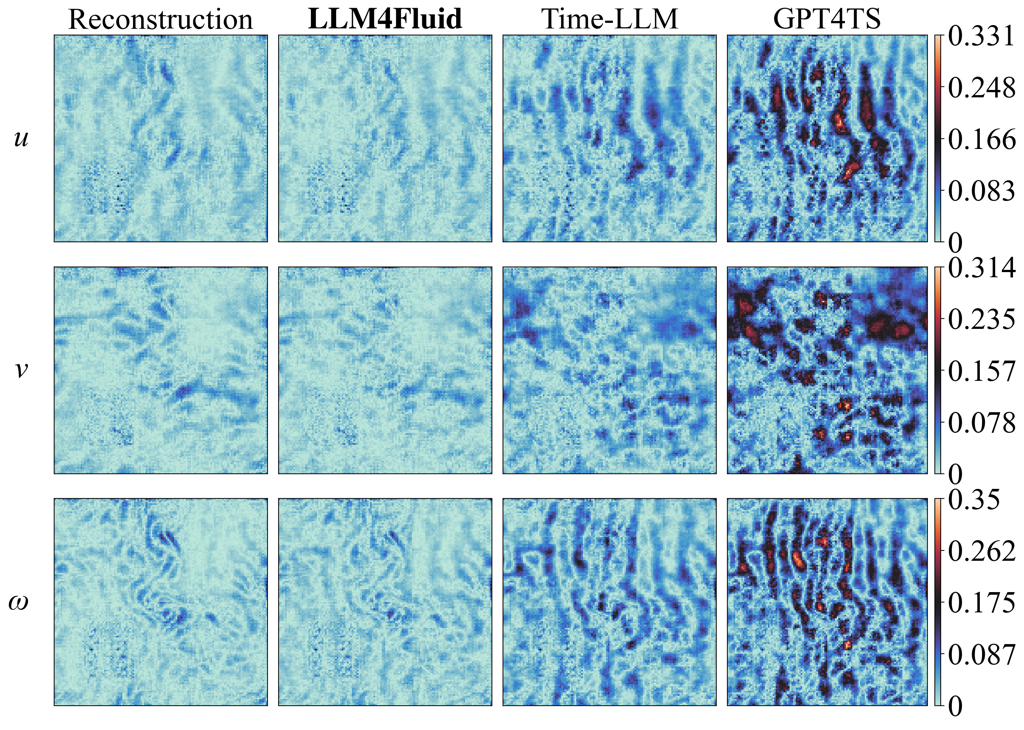}
		\subcaption{}
		\label{fig:10b}
	\end{subfigure}
	
	\caption{Comparison of (a) predicted flow fields and (b) corresponding absolute errors at the final time step for different models on the Low-Re dataset. The predictions of LLM4Fluid are closest to the reconstructed flow fields, achieving the best accuracy.}
	\label{fig:10}
\end{figure*}

\begin{figure*}[t]
	\centering
	\begin{subfigure}[b]{0.49\linewidth}
		\centering
		\includegraphics[width=\linewidth]{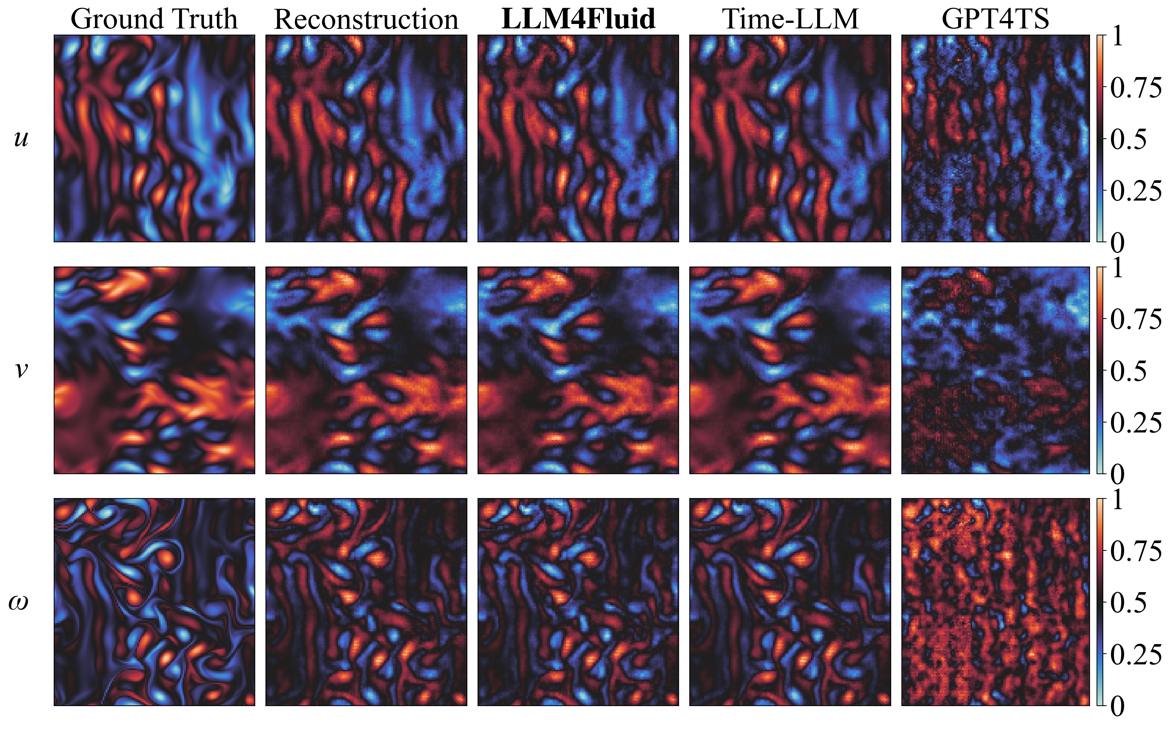}
		\subcaption{}
		\label{fig:11a}
	\end{subfigure}
	\hspace{10pt}%
	\begin{subfigure}[b]{0.427\linewidth}
		\centering
		\includegraphics[width=\linewidth]{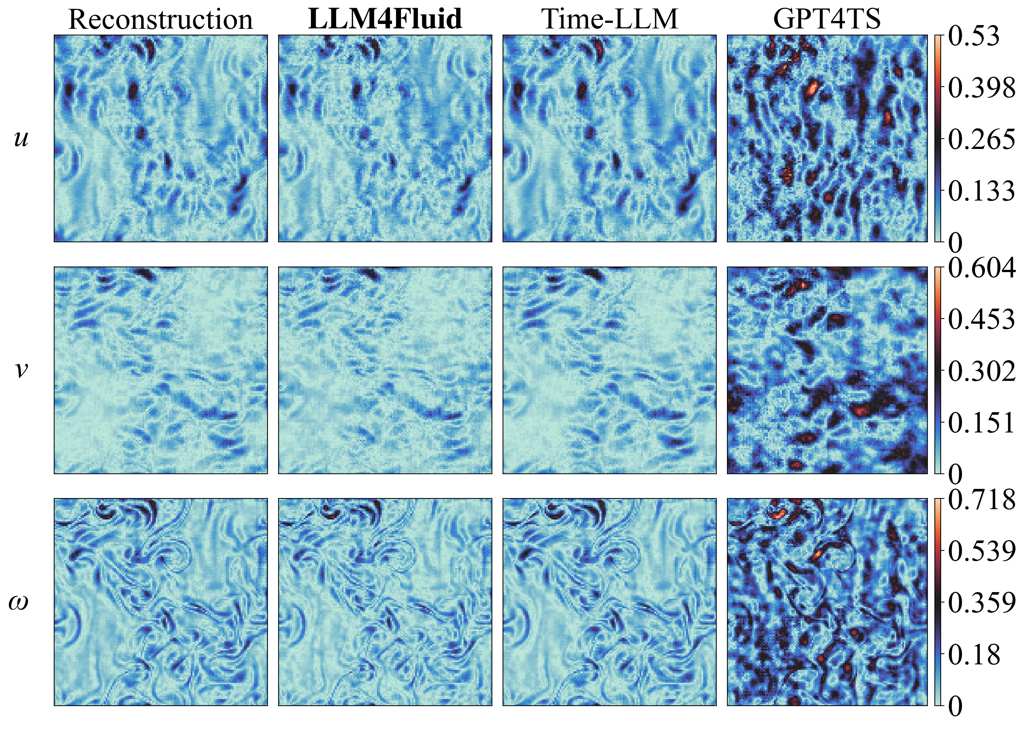}
		\subcaption{}
		\label{fig:11b}
	\end{subfigure}
	
	\caption{Comparison of (a) predicted flow fields and (b) corresponding absolute errors at the final time step for different models on the High-Re dataset. The predictions of LLM4Fluid are closest to the reconstructed flow fields, achieving the best accuracy.}
	\label{fig:11}
\end{figure*}

\begin{figure*}[t]
	\centering
	\begin{subfigure}[b]{0.49\linewidth}
		\centering
		\includegraphics[width=\linewidth]{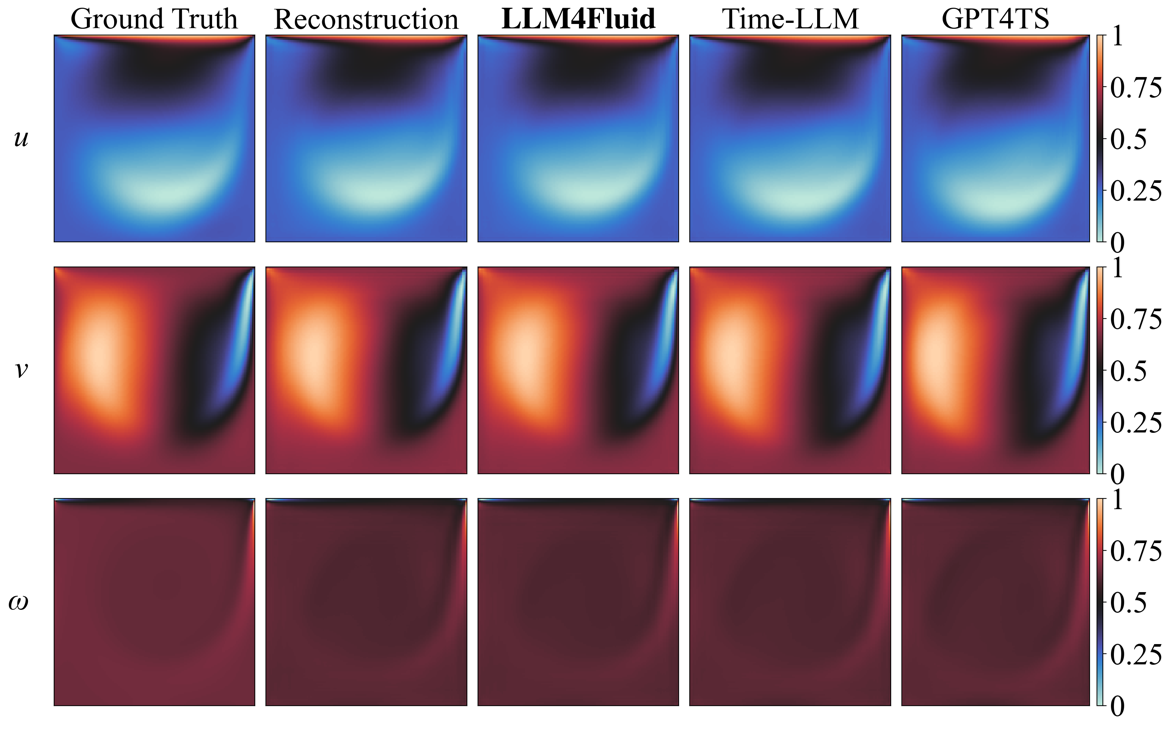}
		\subcaption{}
		\label{fig:12a}
	\end{subfigure}
	\hspace{10pt}%
	\begin{subfigure}[b]{0.427\linewidth}
		\centering
		\includegraphics[width=\linewidth]{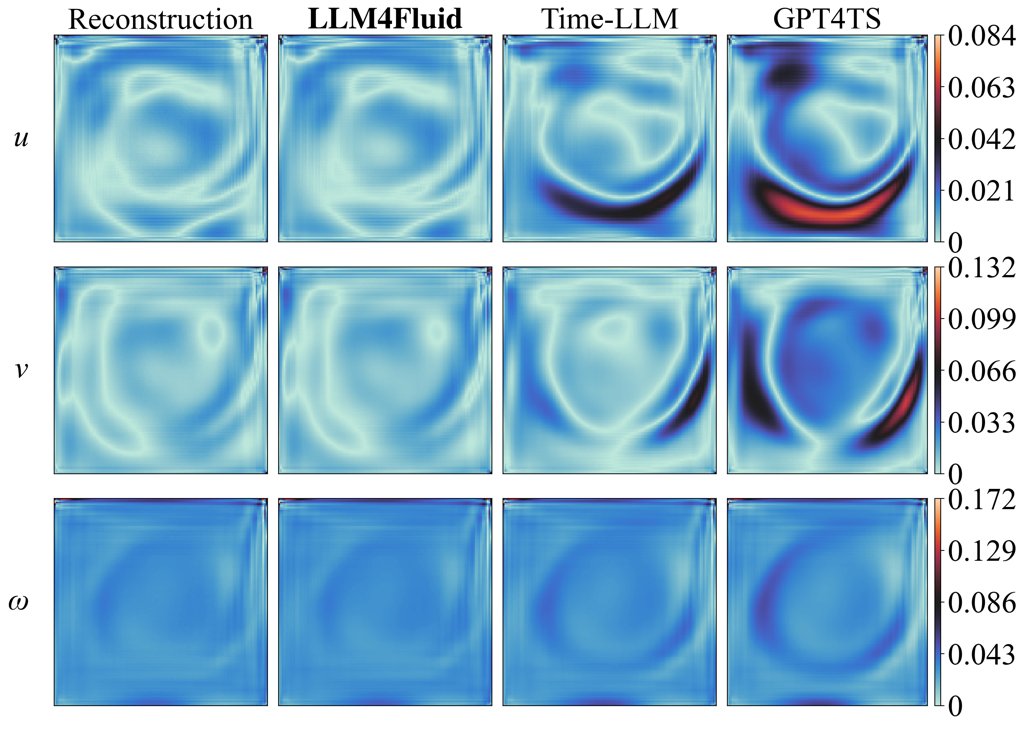}
		\subcaption{}
		\label{fig:12b}
	\end{subfigure}
	
	\caption{Comparison of (a) predicted flow fields and (b) corresponding absolute errors at the final time step for different models on the Cavity dataset. The predictions of LLM4Fluid are closest to the reconstructed flow fields, achieving the best accuracy.}
	\label{fig:12}
\end{figure*}

\begin{figure*}[t]
	\centering
	\begin{subfigure}[b]{0.49\linewidth}
		\centering
		\includegraphics[width=\linewidth]{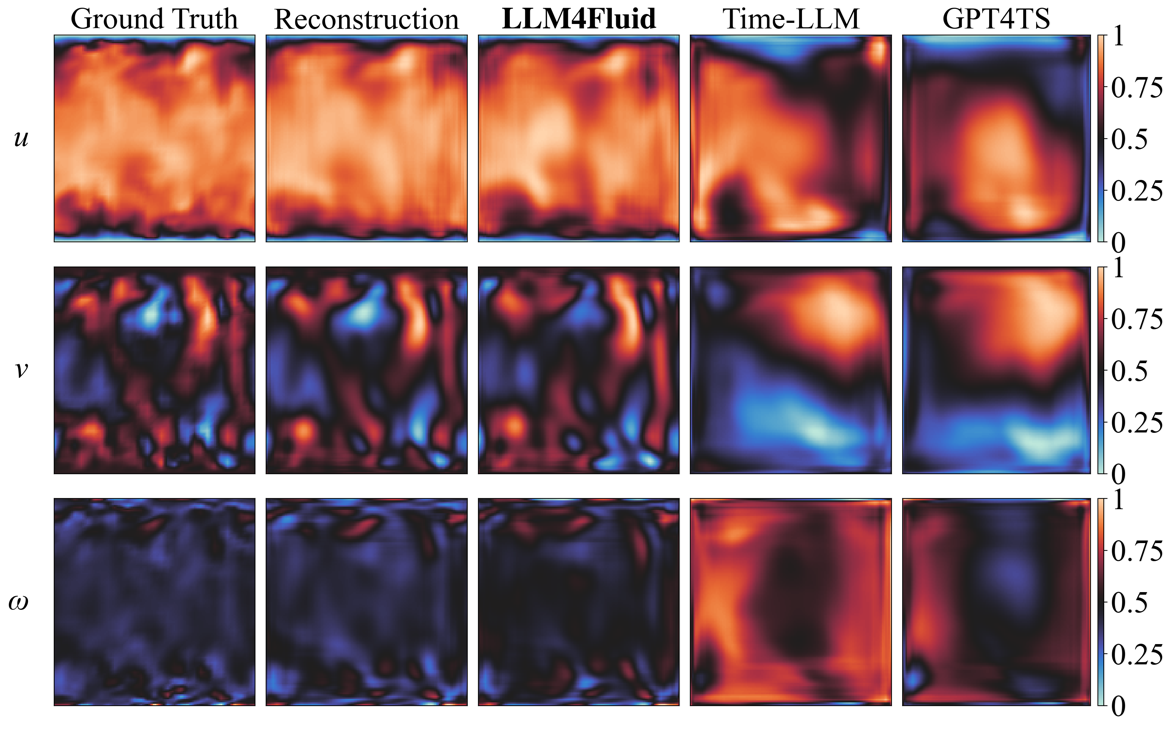}
		\subcaption{}
		\label{fig:13a}
	\end{subfigure}
	\hspace{10pt}%
	\begin{subfigure}[b]{0.427\linewidth}
		\centering
		\includegraphics[width=\linewidth]{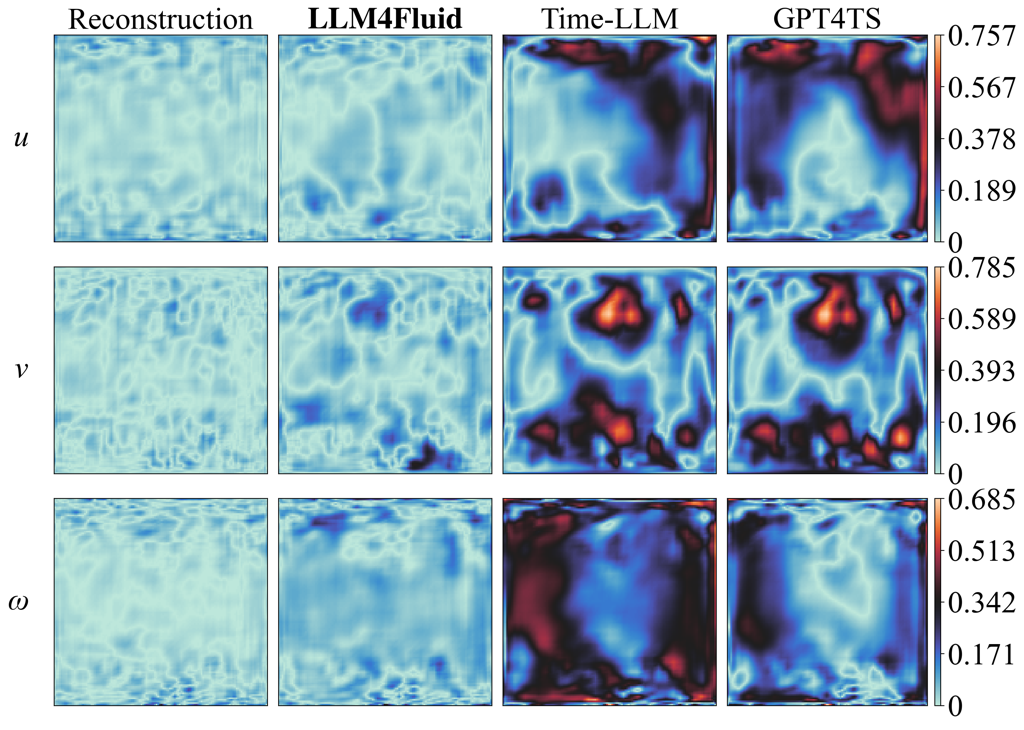}
		\subcaption{}
		\label{fig:13b}
	\end{subfigure}
	
	\caption{Comparison of (a) predicted flow fields and (b) corresponding absolute errors at the final time step for different models on the Channel dataset. The predictions of LLM4Fluid are closest to the reconstructed flow fields, achieving the best accuracy.}
	\label{fig:13}
\end{figure*}

\begin{figure*}[t]
	\centering
	\begin{subfigure}[b]{0.49\linewidth}
		\centering
		\includegraphics[width=\linewidth]{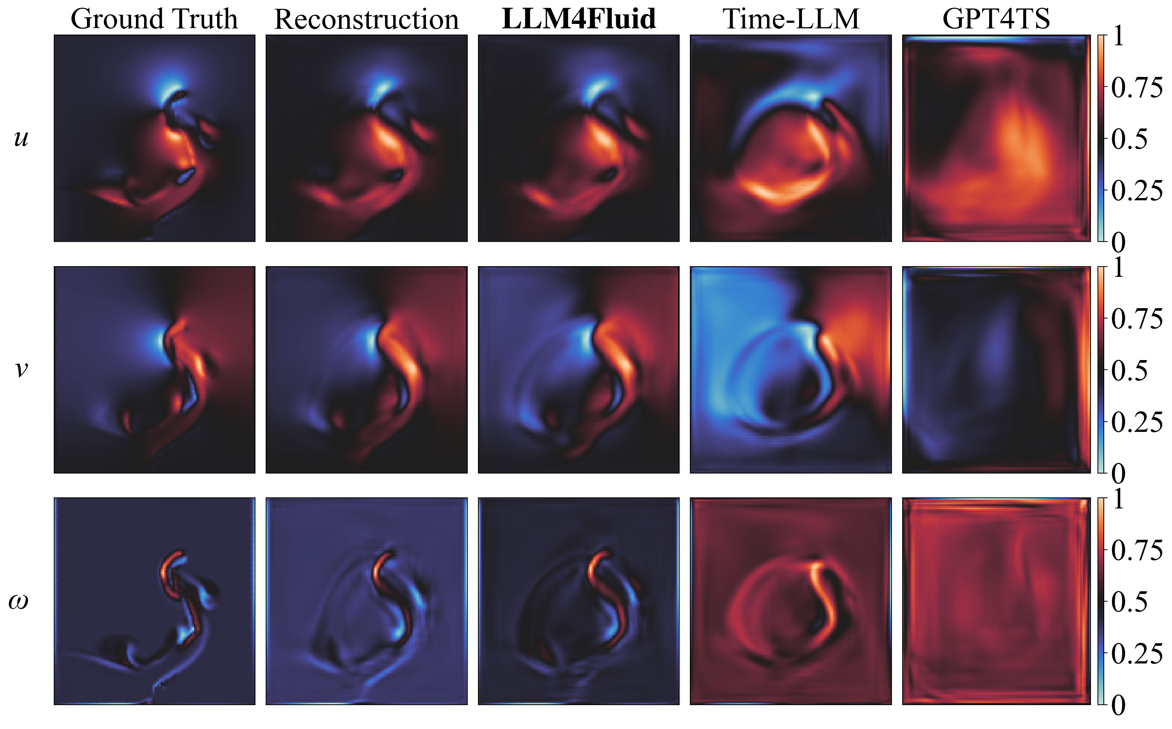}
		\subcaption{}
		\label{fig:14a}
	\end{subfigure}
	\hspace{10pt}%
	\begin{subfigure}[b]{0.427\linewidth}
		\centering
		\includegraphics[width=\linewidth]{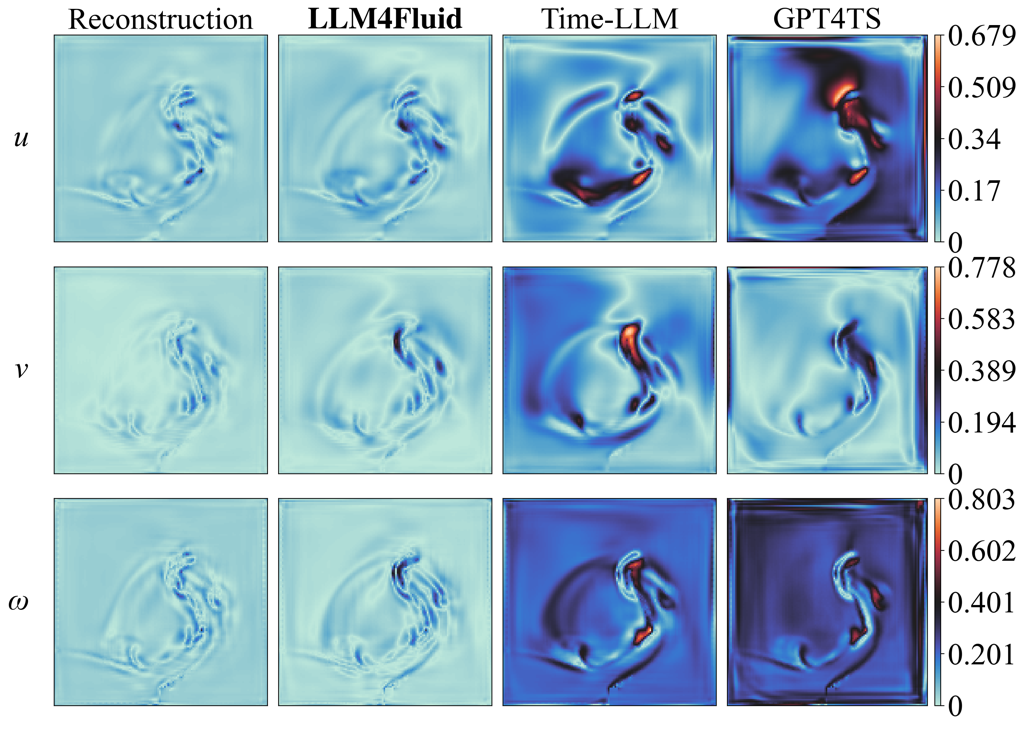}
		\subcaption{}
		\label{fig:14b}
	\end{subfigure}
	
	\caption{Comparison of (a) predicted flow fields and (b) corresponding absolute errors at the final time step for different models on the Dam dataset. The predictions of LLM4Fluid are closest to the reconstructed flow fields, achieving the best accuracy.}
	\label{fig:14}
\end{figure*}

\begin{figure*}[t]
	\centering
	\begin{subfigure}[b]{0.49\linewidth}
		\centering
		\includegraphics[width=\linewidth]{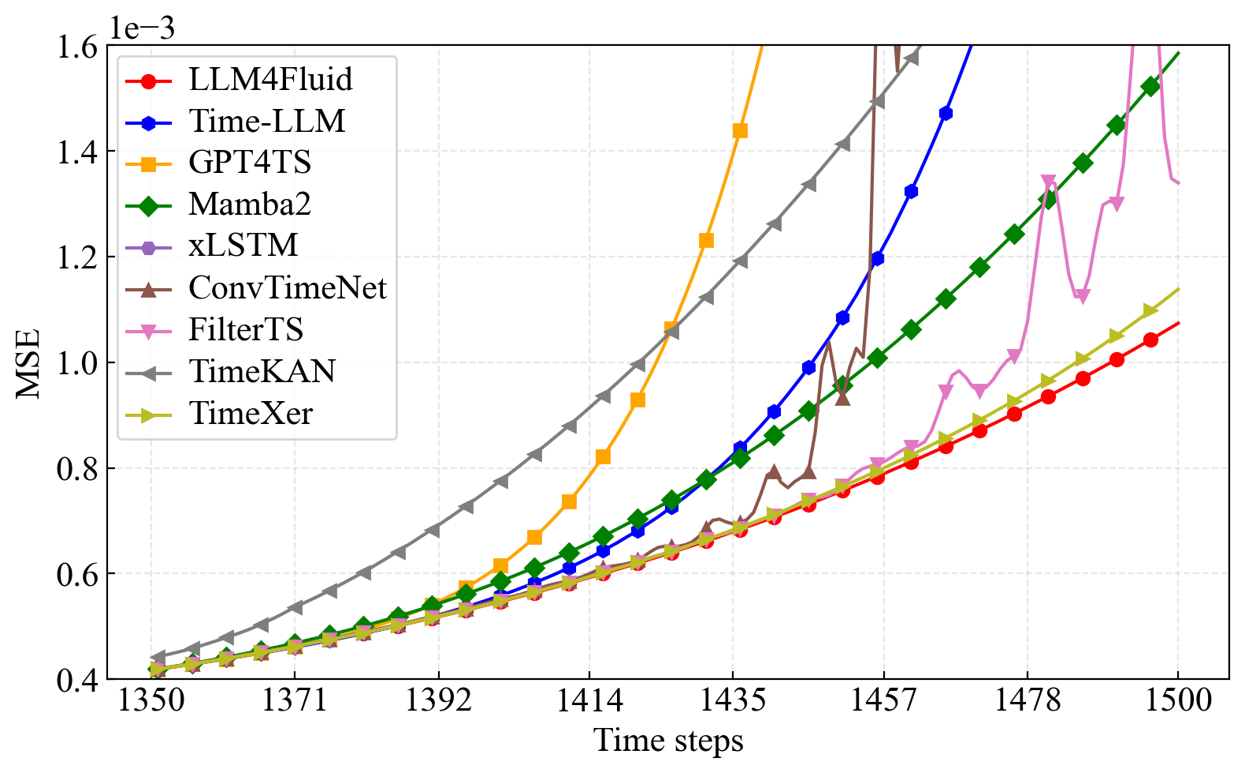}
		\subcaption{}
		\label{fig:15a}
	\end{subfigure}
	\hfill
	\begin{subfigure}[b]{0.49\linewidth}
		\centering
		\includegraphics[width=\linewidth]{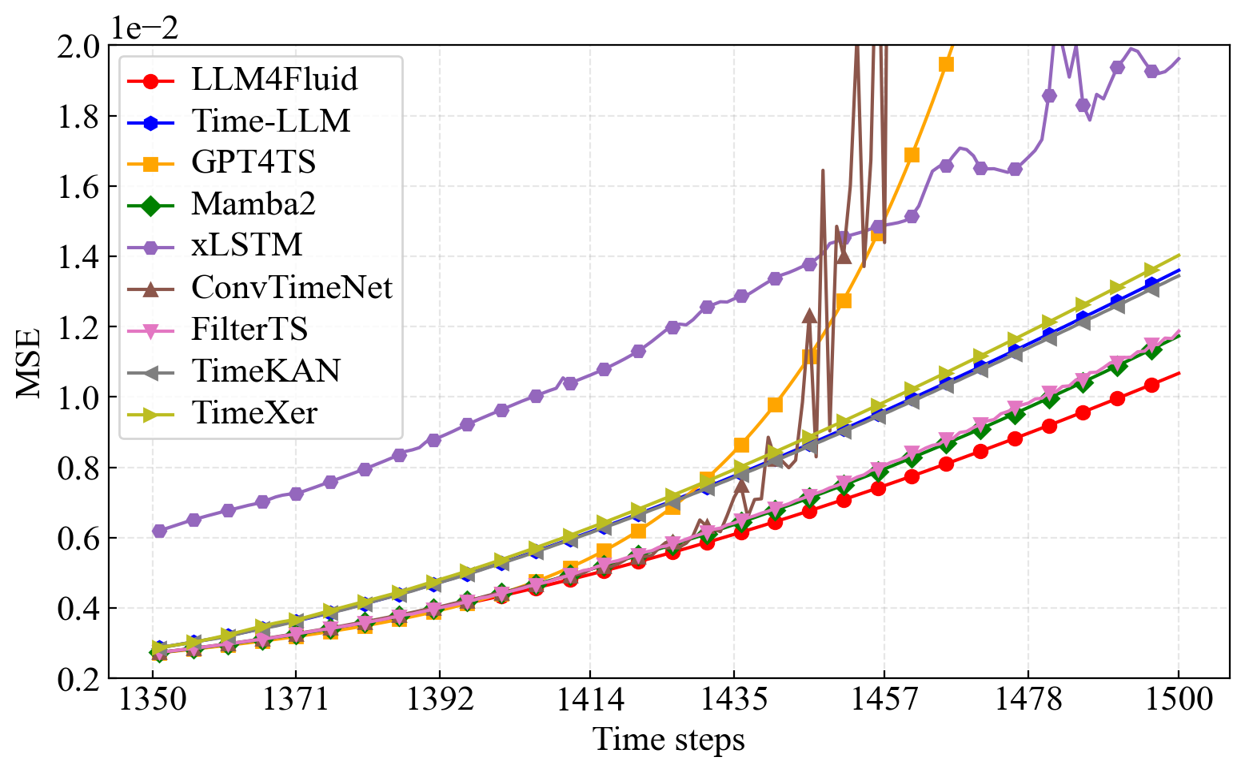}
		\subcaption{}
		\label{fig:15b}
	\end{subfigure}
	
	\begin{subfigure}[b]{0.49\linewidth}
		\centering
		\includegraphics[width=\linewidth]{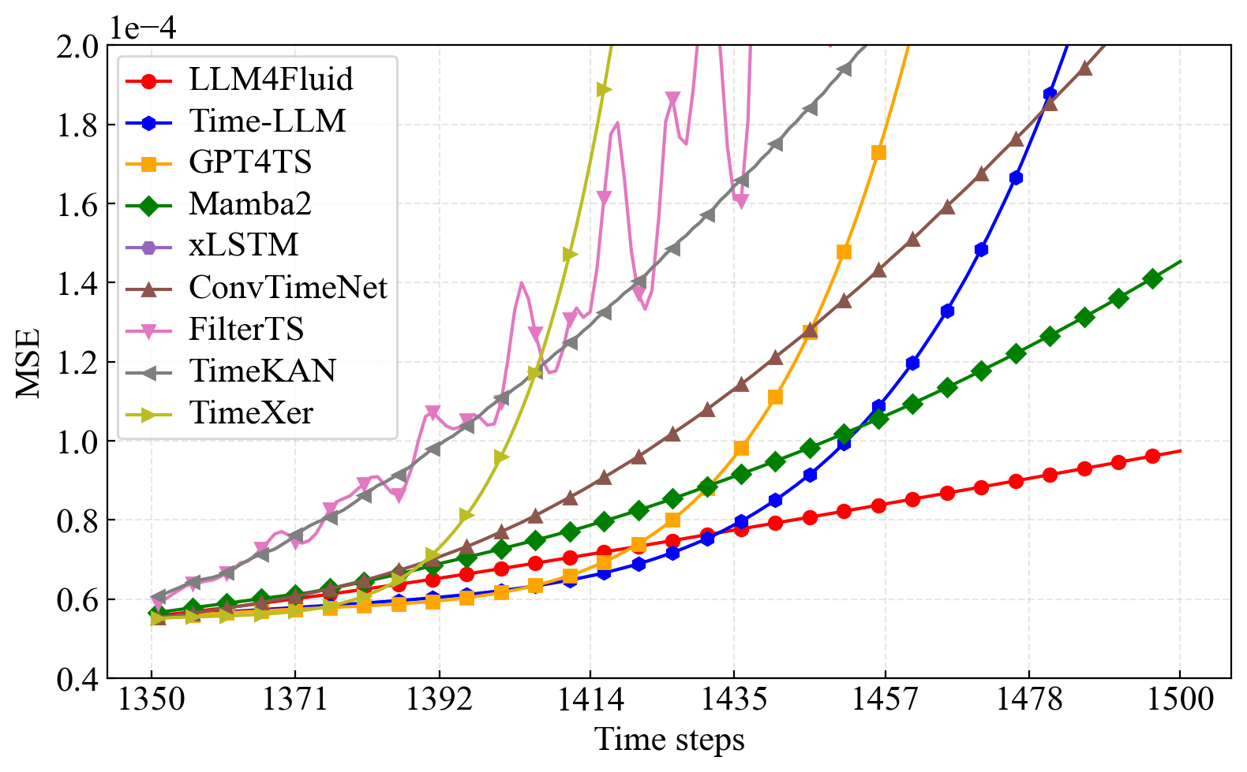}
		\subcaption{}
		\label{fig:15c}
	\end{subfigure}
	\hfill
	\begin{subfigure}[b]{0.49\linewidth}
		\centering
		\includegraphics[width=\linewidth]{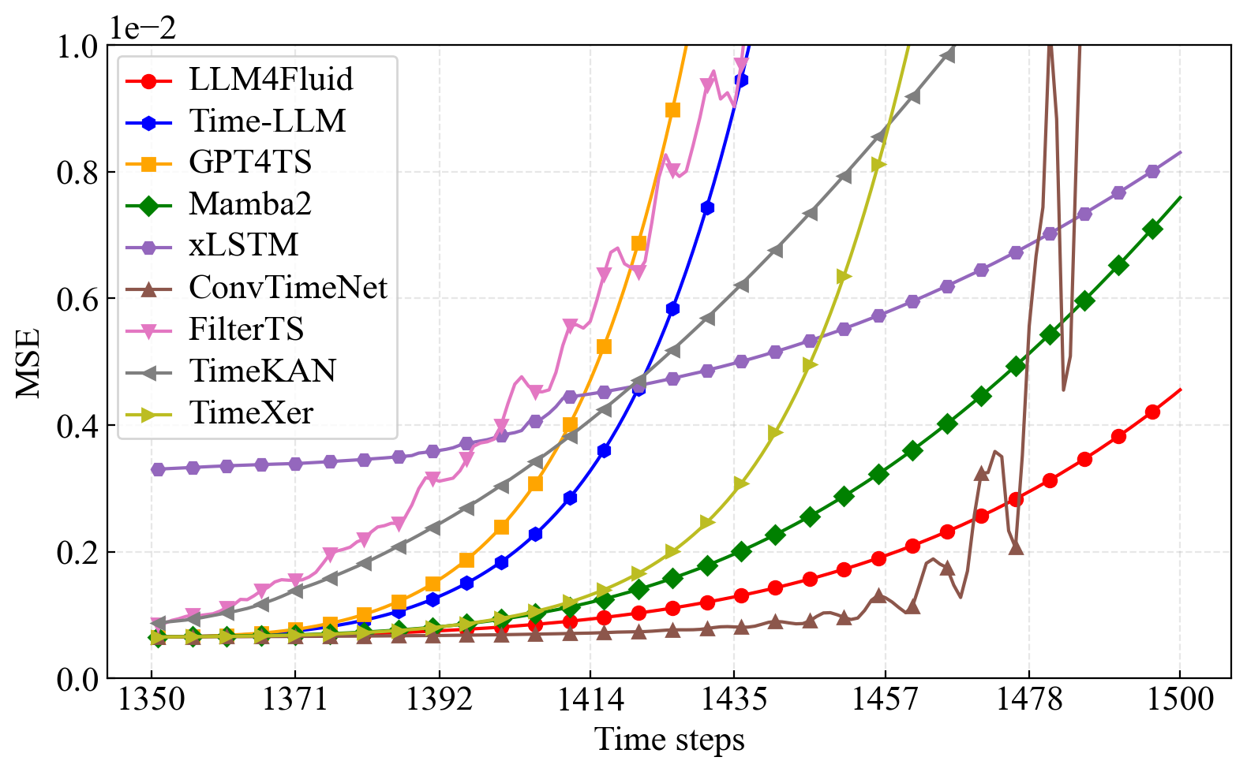}
		\subcaption{}
		\label{fig:15d}
	\end{subfigure}

	\begin{subfigure}[b]{0.49\linewidth}
		\centering
		\includegraphics[width=\linewidth]{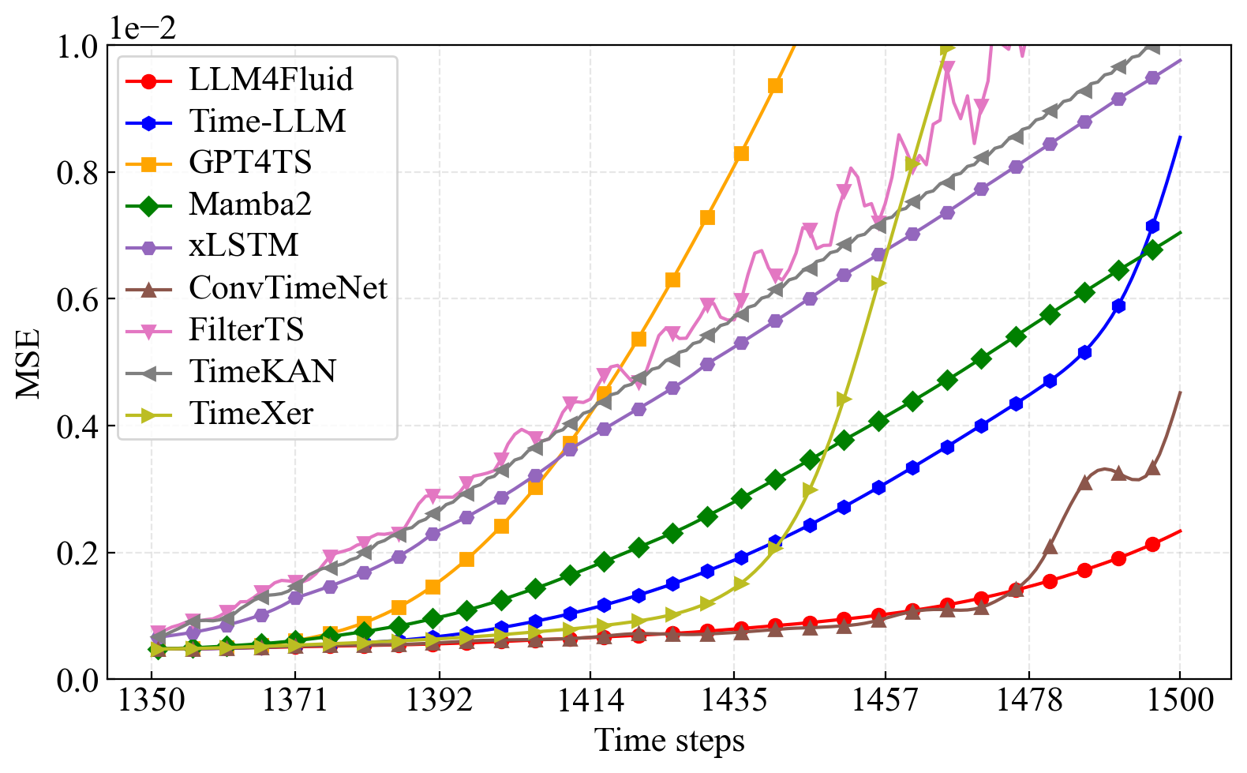}
		\subcaption{}
		\label{fig:15e}
	\end{subfigure}

	\caption{Temporal evolution of MSE for different models on the (a) Low-Re, (b) High-Re, (c) Cavity, (d) Channel, and (e) Dam datasets. LLM4Fluid exhibits the slowest error accumulation over time compared with other baselines.}
	\label{fig:15}
\end{figure*}

\begin{figure*}[t]
	\centering
	\begin{subfigure}[b]{0.48\linewidth}
		\centering
		\includegraphics[width=\linewidth]{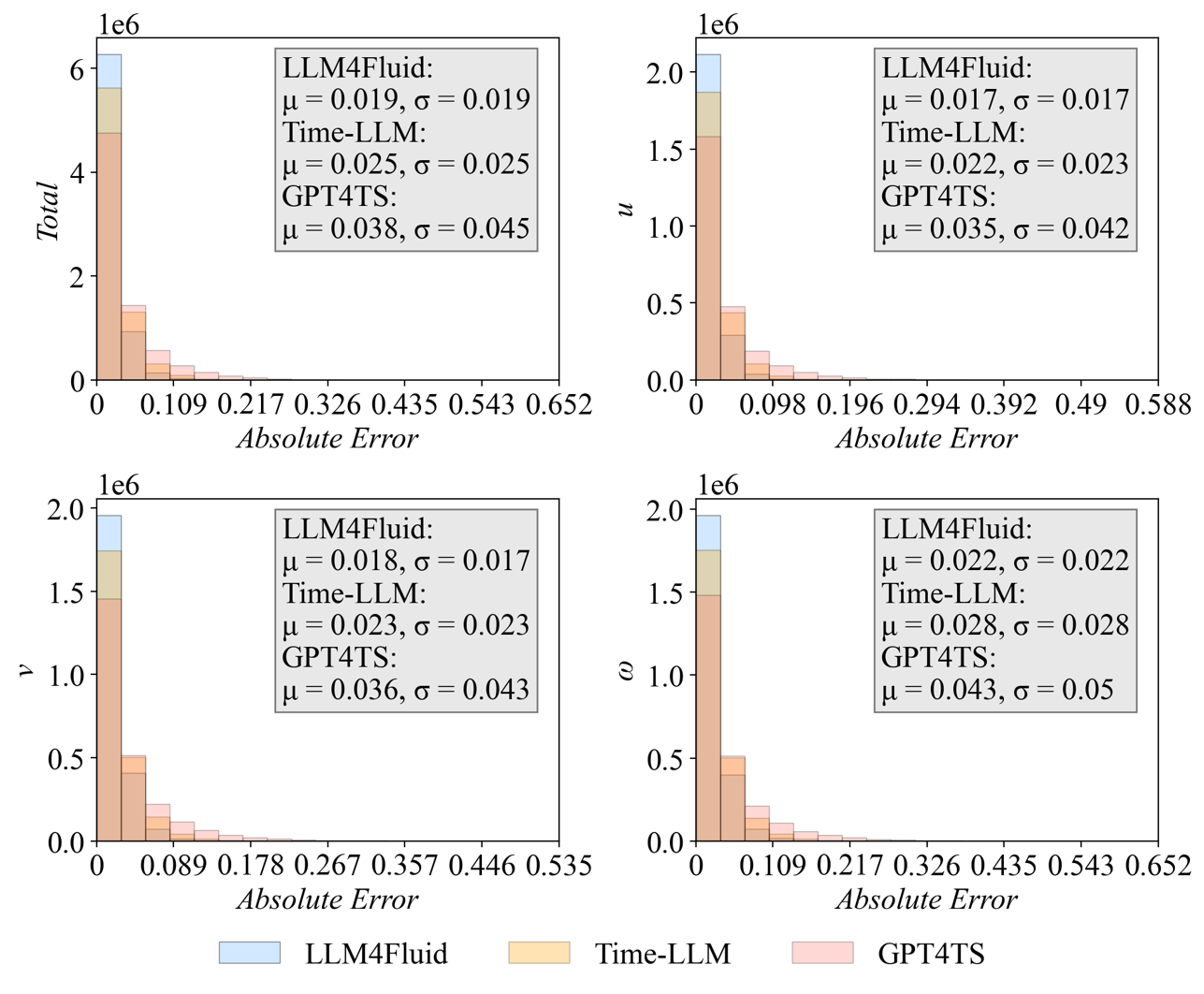}
		\subcaption{}
		\label{fig:16a}
	\end{subfigure}
	\hfill
	\begin{subfigure}[b]{0.48\linewidth}
		\centering
		\includegraphics[width=\linewidth]{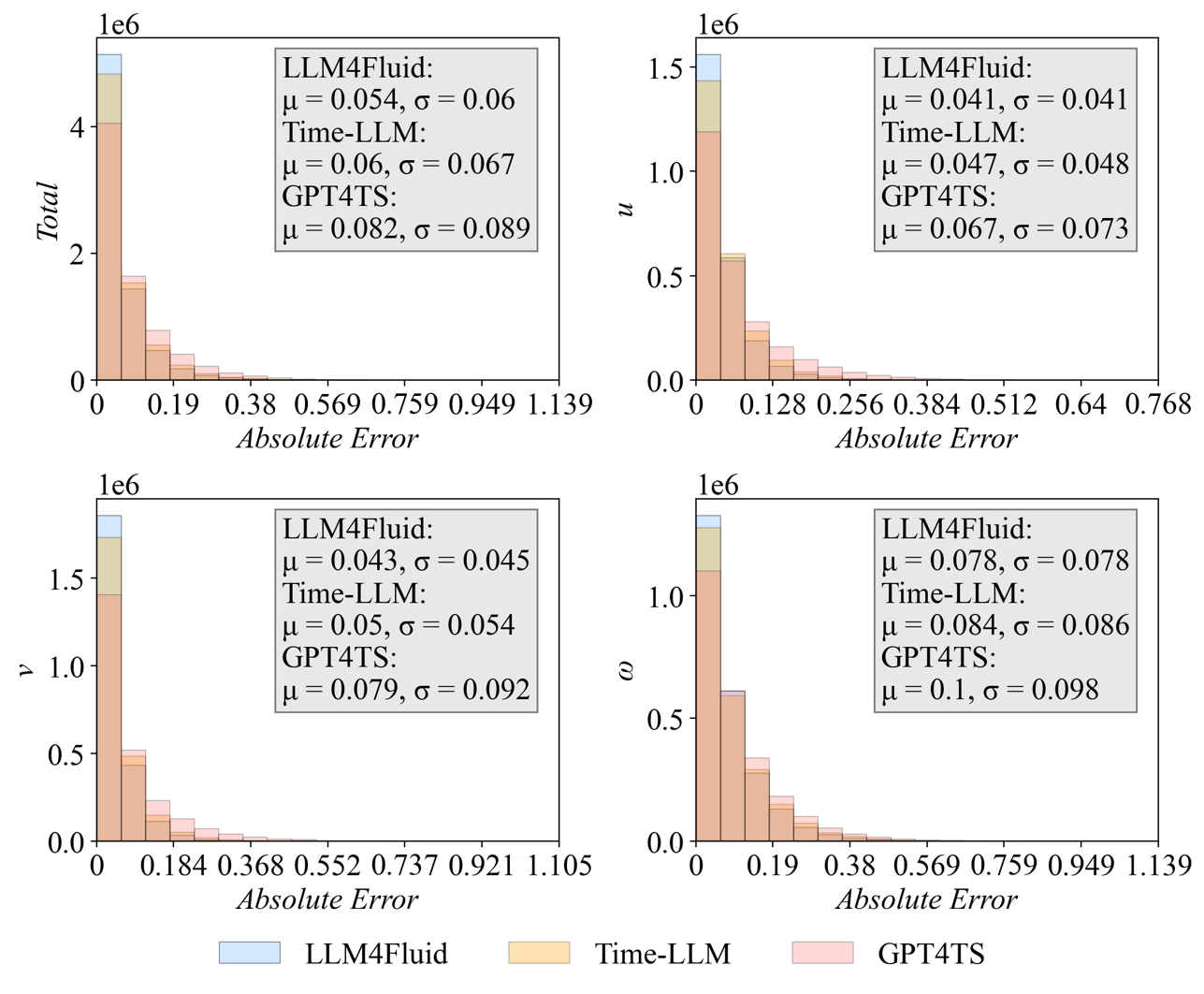}
		\subcaption{}
		\label{fig:16b}
	\end{subfigure}
	
	\begin{subfigure}[b]{0.48\linewidth}
		\centering
		\includegraphics[width=\linewidth]{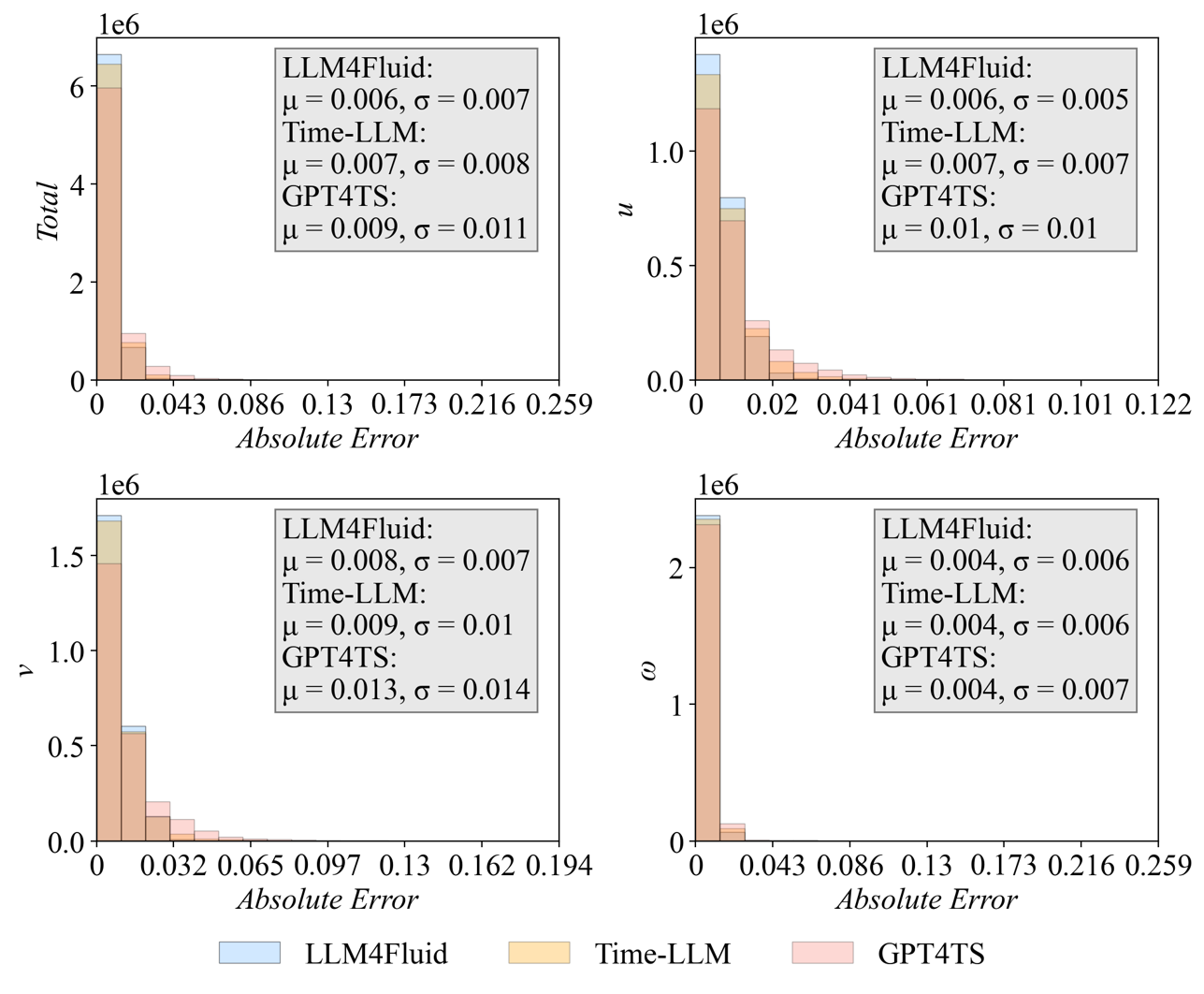}
		\subcaption{}
		\label{fig:16c}
	\end{subfigure}
	\hfill
	\begin{subfigure}[b]{0.48\linewidth}
		\centering
		\includegraphics[width=\linewidth]{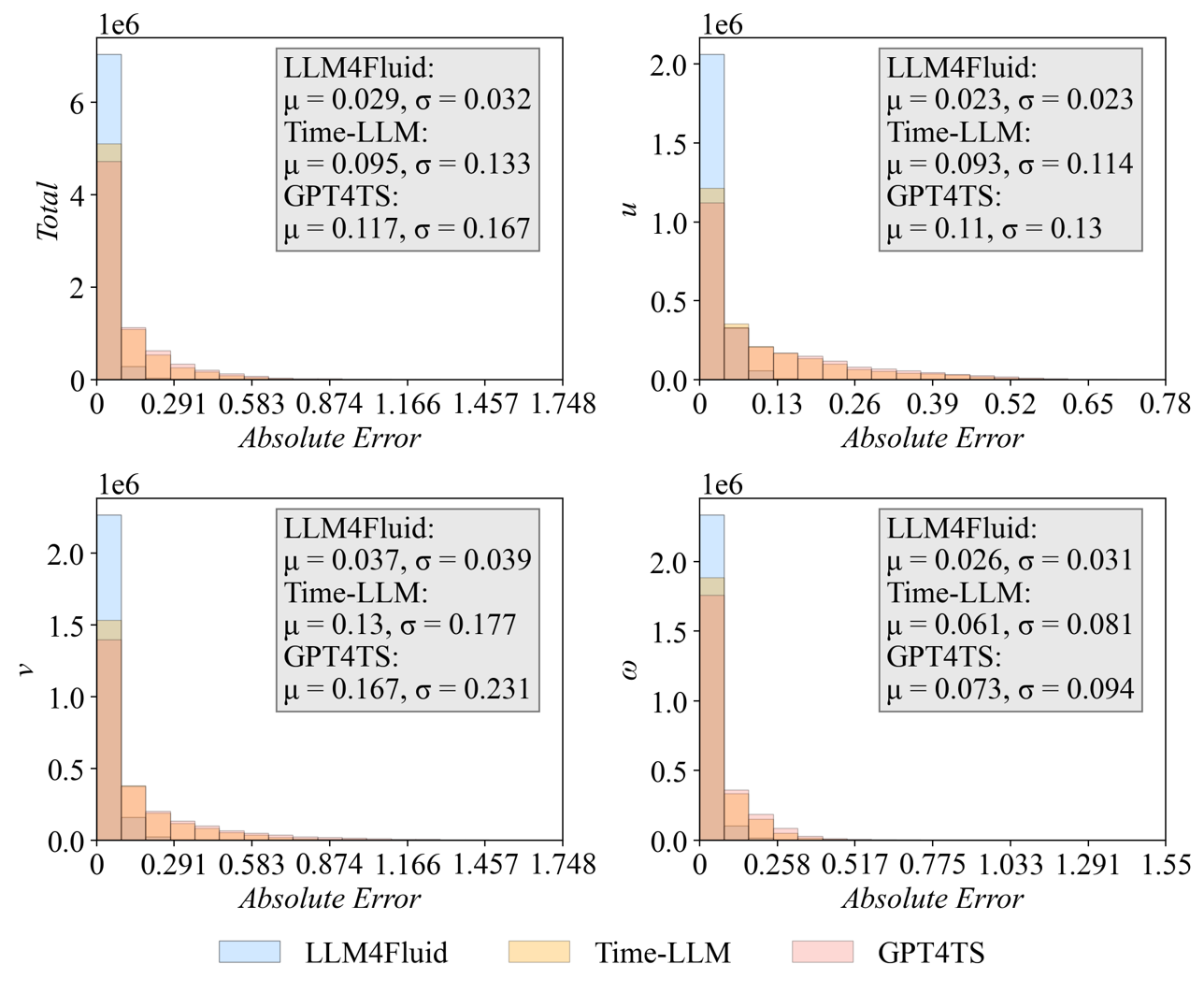}
		\subcaption{}
		\label{fig:16d}
	\end{subfigure}
	
	\begin{subfigure}[b]{0.48\linewidth}
		\centering
		\includegraphics[width=\linewidth]{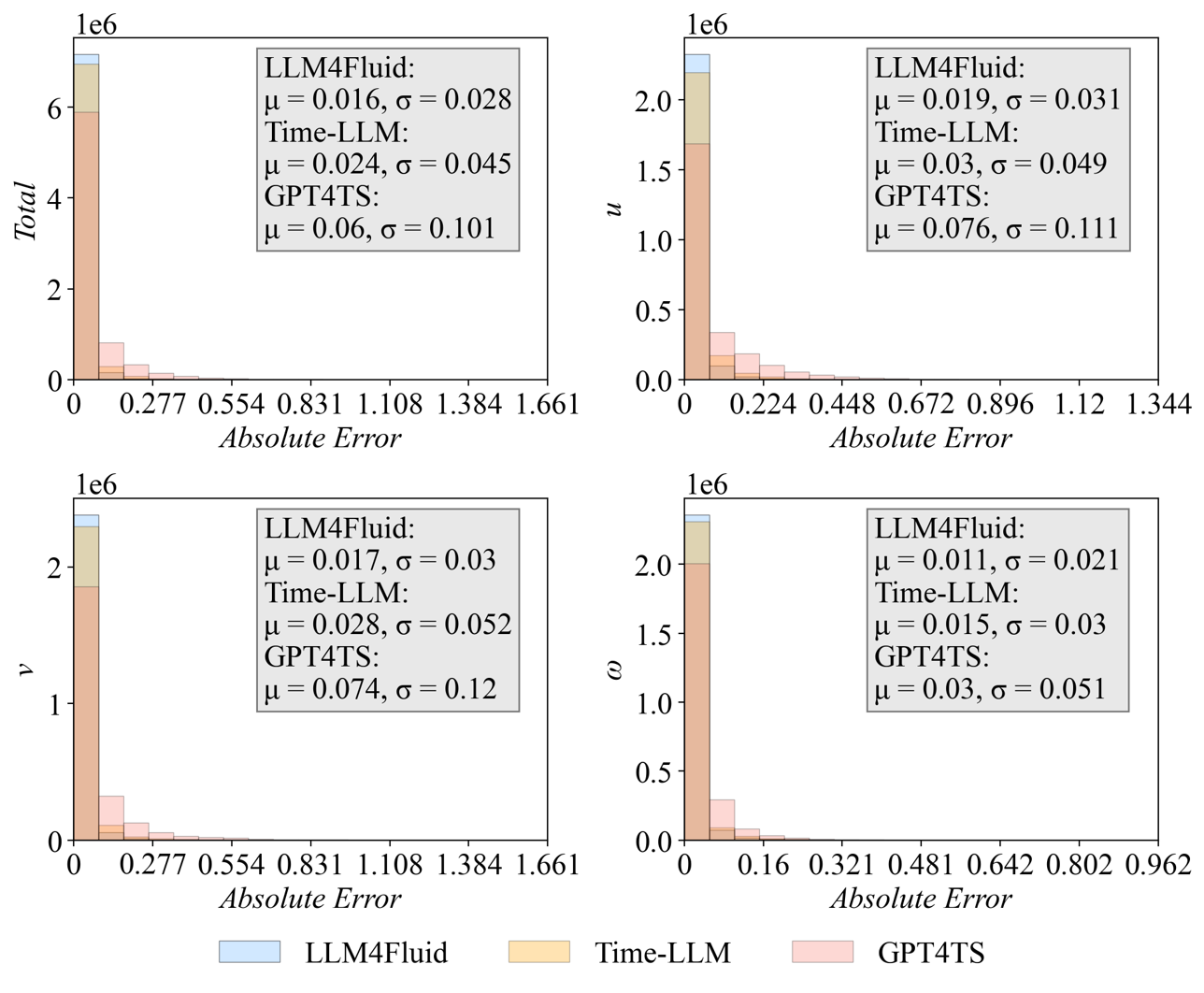}
		\subcaption{}
		\label{fig:16e}
	\end{subfigure}
	
	\caption{Distribution of absolute errors for physical variables on the (a) Low-Re, (b) High-Re, (c) Cavity, (d) Channel, and (e) Dam datasets. \textit{Total} represents the combined error distribution of \textit{u}, \textit{v}, and $\omega$. The $\mu$ and $\sigma$ denote the mean and standard deviation of absolute errors, respectively. LLM4Fluid consistently achieves smaller values in both metrics, reflecting better accuracy and stability.}
	\label{fig:16}
\end{figure*}

\begin{figure*}[t]
	\centering
	\centering
	\includegraphics[width=0.95\linewidth]{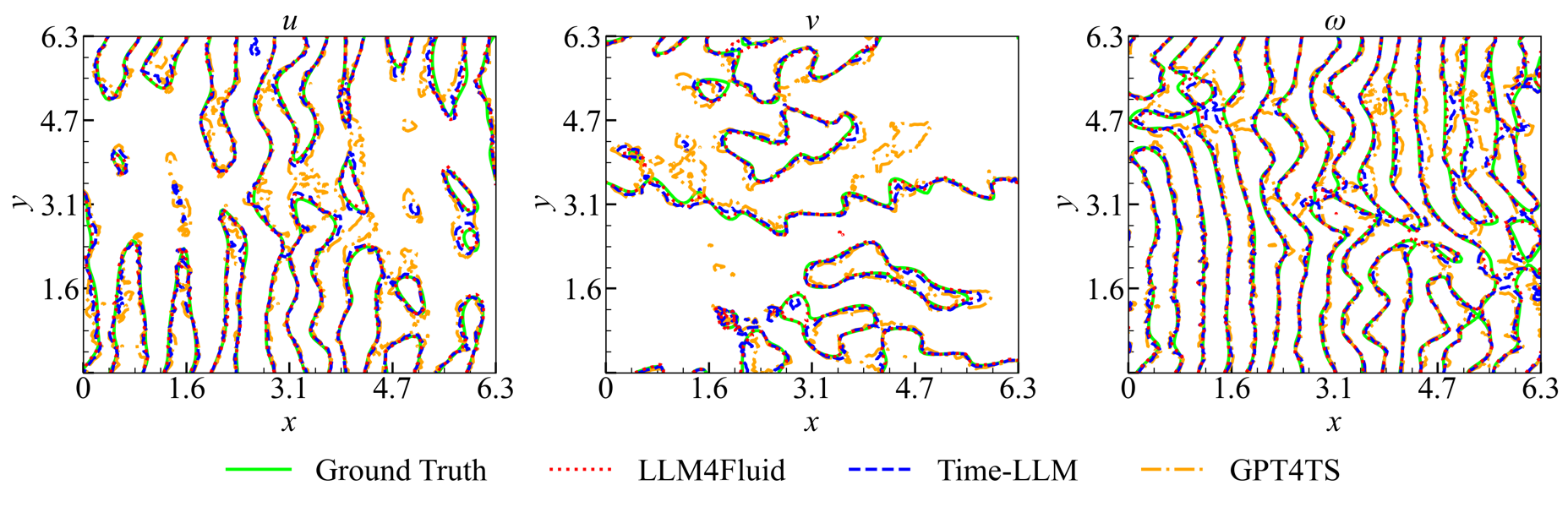}
	
	\caption{Comparison of contour lines of flow fields between the ground truth and different prediction models on the Low-Re dataset at the final time step. LLM4Fluid maintains the closest match to the ground truth.}
	\label{fig:17}
\end{figure*}

\begin{figure*}[t]
	\centering
	\includegraphics[width=0.95\linewidth]{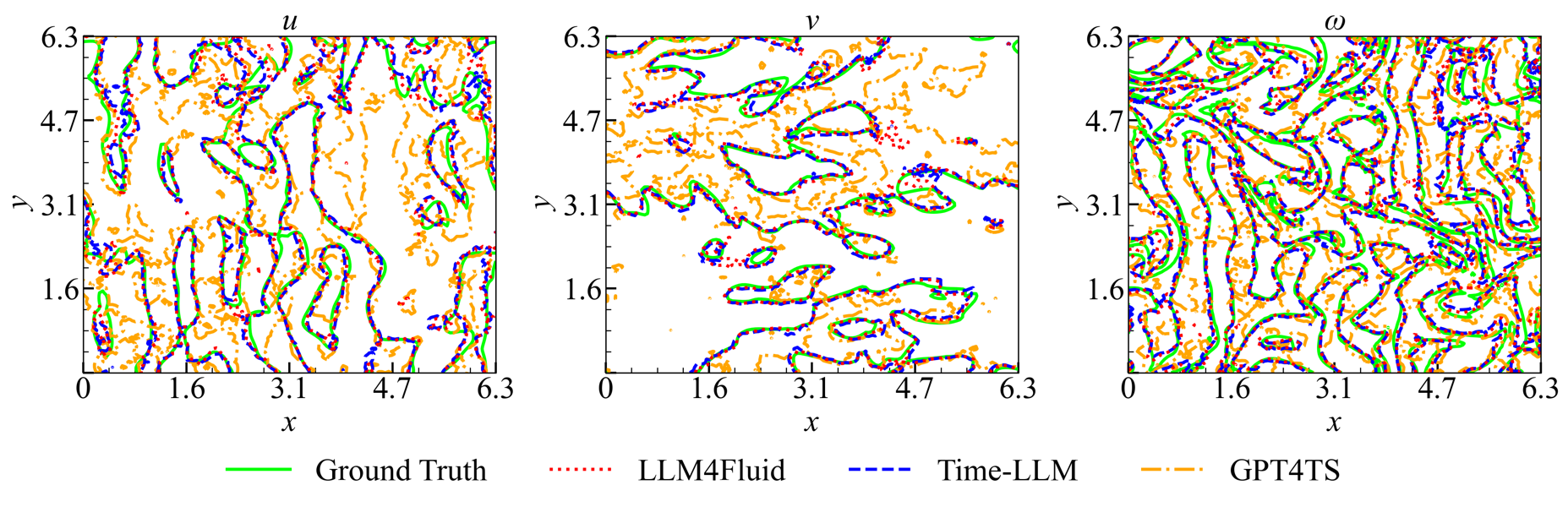}
	
	\caption{Comparison of contour lines of flow fields between the ground truth and different prediction models on the High-Re dataset at the final time step. LLM4Fluid maintains the closest match to the ground truth.}
	\label{fig:18}
\end{figure*}

\begin{figure*}[t]
	\centering
	\includegraphics[width=0.95\linewidth]{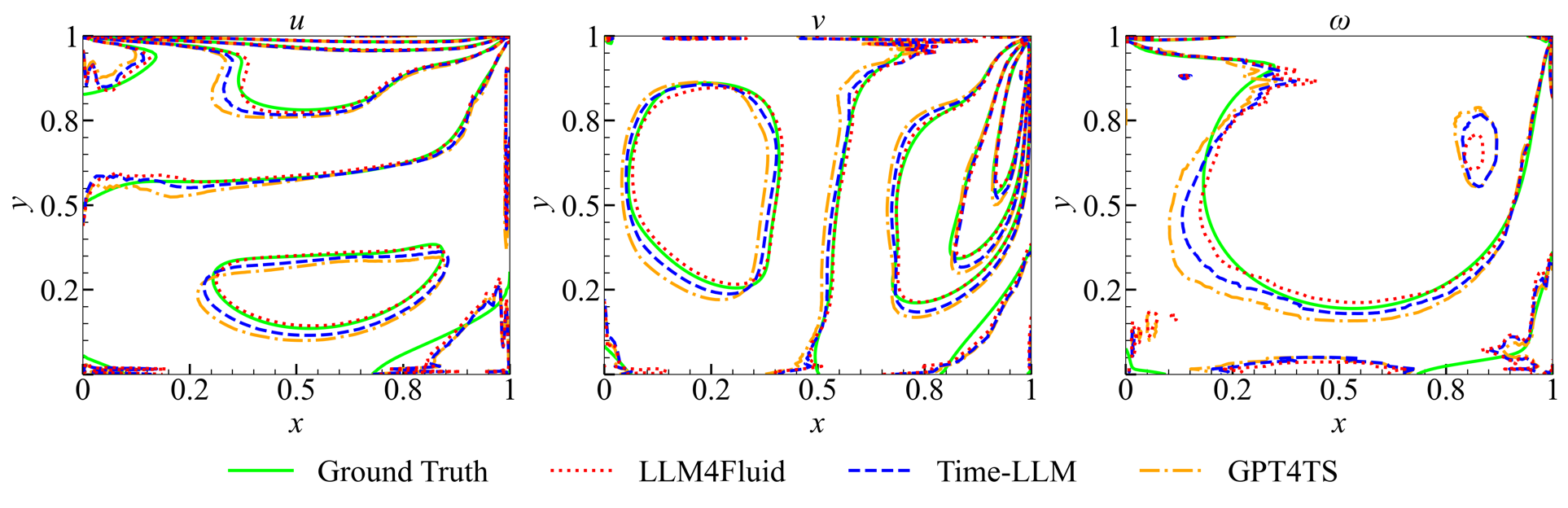}
	
	\caption{Comparison of contour lines of flow fields between the ground truth and different prediction models on the Cavity dataset at the final time step. LLM4Fluid maintains the closest match to the ground truth.}
	\label{fig:19}
\end{figure*}

\begin{figure*}[t]
	\centering
	\includegraphics[width=0.95\linewidth]{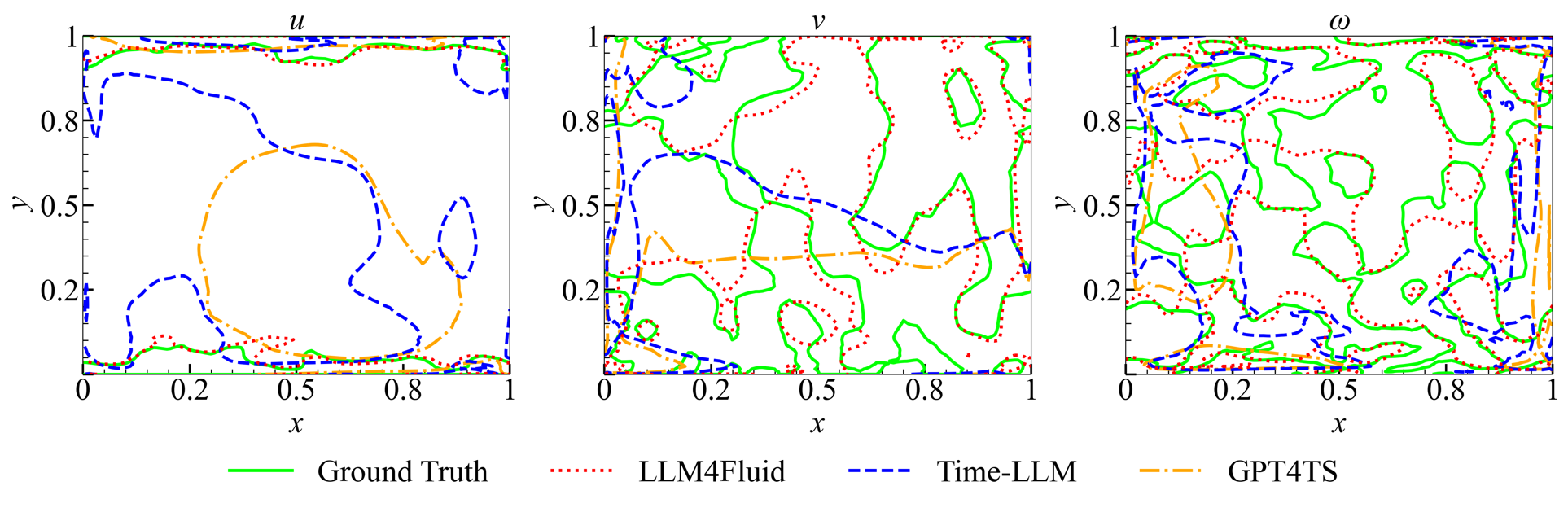}
	
	\caption{Comparison of contour lines of flow fields between the ground truth and different prediction models on the Channel dataset at the final time step. LLM4Fluid maintains the closest match to the ground truth.}
	\label{fig:20}
\end{figure*}

\begin{figure*}[t]
	\centering
	\includegraphics[width=0.95\linewidth]{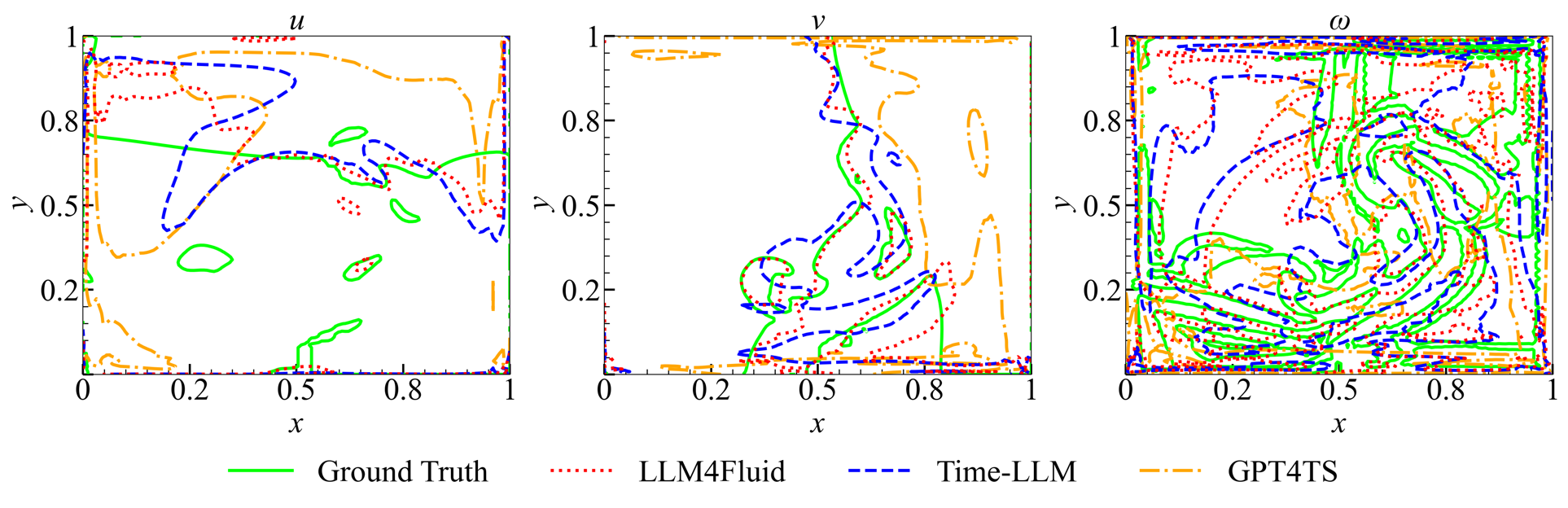}
	
	\caption{Comparison of contour lines of flow fields between the ground truth and predictions of  different models on the Dam dataset at the final time step. LLM4Fluid maintains the closest match to the ground truth.}
	\label{fig:21}
\end{figure*}

\subsection{Latent Space Analysis} \label{sec:latent space analysis}

To further analyze the latent space of the proposed disentangled reduced-order modeling method, we first conduct quantitative ablation studies on the latent dimension. As shown in \cref{tab:6}, varying the latent dimension from 8 to 128 reveals a clear trade-off between compression capacity and redundancy. Too few latent variables (e.g., $D$=8) lead to noticeable information loss and higher reconstruction errors, while excessively large dimensions (e.g., $D$=128) introduce redundant and potentially entangled features that may even hinder reconstruction accuracy. A moderate latent size ($D$=32) yields the best reconstruction performance, indicating that it provides sufficient expressive power while keeping the latent representation compact and stable.

Beyond the quantitative results, we also visualize the latent space using t-SNE. Specifically, we project the latent representations from five datasets onto a two-dimensional plane, as illustrated in \cref{fig:9}. The resulting embeddings form five distinct and smooth trajectories, each corresponding to a different flow scenario. The clear separation between these manifolds indicates that the learned latent space captures physically meaningful variations and successfully distinguishes flow behaviors across diverse scenarios. These results demonstrate that the proposed disentangled reduced-order modeling method yields a compact, well-structured, and physics-aware latent space that is well-suited for downstream temporal modeling.

\subsection{Flow Field Prediction} \label{sec:flow field prediction}

We evaluate LLM4Fluid with a series of LLM backbones of varying sizes, including GPT-2 and the OPT family from 125M to 6.7B parameters, as summarized in \cref{tab:7}. Larger backbones consistently yield lower errors, with OPT-6.7B achieving the best overall performance. These results indicate that LLM4Fluid can effectively leverage the additional capacity of larger language models, while also delivering competitive accuracy when instantiated with smaller, more efficient backbones.

Comparisons of the predicted flow fields and the corresponding absolute error maps at the final time step are shown in \cref{fig:10,fig:11,fig:12,fig:13,fig:14} for the Low-Re, High-Re, Cavity, Channel, and Dam datasets, respectively. Across all scenarios, LLM4Fluid produces predictions that most closely match the ground truth and exhibits the smallest absolute errors, demonstrating its strong capability for spatio-temporal flow field prediction.

\cref{fig:15} compares the temporal evolution of MSE for different models on five datasets. Most methods start from comparable error levels at the beginning of the rollout, but the baselines quickly accumulate errors as time advances, leading to a much steeper growth of MSE. In contrast, LLM4Fluid maintains the lowest MSE throughout the prediction horizon and exhibits the slowest error growth on all datasets, indicating significantly reduced error accumulation. These results demonstrate that the proposed disentangled reduced-order modeling method enhanced with a physics-informed disentanglement mechanism, and the LLM-based temporal processor with a modality alignment strategy, together provide more stable long-term predictions of fluid dynamics.

\cref{fig:16} presents the histograms of absolute errors for the total field and individual variables \textit{u}, \textit{v}, and $\omega$ on five datasets. For all datasets, the error distributions of LLM4Fluid are more sharply concentrated near zero and exhibit lighter tails than those of Time-LLM and GPT4TS. This is also reflected in the reported statistics: LLM4Fluid attains the smallest mean $\mu$ and standard deviation $\sigma$ of absolute errors across all variables, indicating not only higher overall prediction accuracy but also improved stability with fewer large local errors in the predicted flow fields.

We compare the contour lines of flow fields between the ground truth and different prediction models at the final time step across five datasets, as illustrated in \cref{fig:17,fig:18,fig:19,fig:20,fig:21}. In all cases, the contours produced by LLM4Fluid (red dotted lines) almost overlap with the ground-truth curves (green solid lines), indicating that the model accurately captures the spatial distribution and phase of coherent structures such as shear layers and vortical filaments. Time-LLM (blue dashed lines) also follows the true contours reasonably well but exhibits local deviations, especially near regions with strong gradients. GPT4TS (orange dash–dot lines) shows the largest discrepancies, with distorted or shifted contours and missing fine-scale structures. These results highlight that LLM4Fluid maintains the closest match to the ground truth, demonstrating superior fidelity in predicting detailed flow patterns under complex flow scenarios.


\end{document}